\documentclass{article}
\usepackage[usenames,dvipsnames,svgnames,table]{xcolor}
\usepackage[preprint,nonatbib]{neurips_2024}
\usepackage[table, dvipsnames]{xcolor} 
\usepackage{caption}
\usepackage[utf8]{inputenc} 
\usepackage[T1]{fontenc}    
\usepackage{hyperref}       
\usepackage{url}            
\usepackage{booktabs}       
\usepackage{amsfonts}       
\usepackage{nicefrac}       
\usepackage{microtype}      
\usepackage{xcolor}         
\usepackage[numbers,comma,sort]{natbib}
\usepackage{amsmath,amsfonts,amsthm,bm}
\usepackage{graphicx}
\usepackage{booktabs}
\usepackage{amssymb}
\usepackage{multirow}
\usepackage{algorithm}
\usepackage{algpseudocode}
\title{CryoGEM: Physics-Informed Generative Cryo-Electron Microscopy}
\newcommand*\samethanks[1][\value{footnote}]{\footnotemark[#1]}
\vspace{-10pt}
\author{Jiakai Zhang$^{1,2\thanks{The authors contributed equally to this work.}}$ ~~ Qihe Chen$^{1,2\samethanks[1]}$ ~~ Yan Zeng$^{1,2}$ ~~ Wenyuan Gao$^{1}$ \AND Xuming He$^{1}$  ~~ Zhijie Liu$^{1,3}$ ~~ Jingyi Yu$^{1}$\\ \\$^1$ShanghaiTech University\\$^2$Cellverse \\$^3$iHuman Institute \\\\{\tt\small \{zhangjk,chenqh2024,zengyan2024,gaowy,hexm,liuzhj,yujingyi\}@shanghaitech.edu.cn}}
\usepackage{xspace}

\newcommand{\ours}{CryoGEM\xspace}

\begin{document}
\maketitle
\vspace{-10pt}
\begin{abstract}
In the past decade, deep conditional generative models have revolutionized the generation of realistic images, extending their application from entertainment to scientific domains. Single-particle cryo-electron microscopy (cryo-EM) is crucial in resolving near-atomic resolution 3D structures of proteins, such as the SARS-COV-2 spike protein. To achieve high-resolution reconstruction, a comprehensive data processing pipeline has been adopted. However, its performance is still limited as it lacks high-quality annotated datasets for training. To address this, we introduce physics-informed generative cryo-electron microscopy (CryoGEM), which for the first time integrates physics-based cryo-EM simulation with a generative unpaired noise translation to generate physically correct synthetic cryo-EM datasets with realistic noises. Initially, CryoGEM simulates the cryo-EM imaging process based on a virtual specimen. To generate realistic noises, we leverage an unpaired noise translation via contrastive learning with a novel mask-guided sampling scheme. Extensive experiments show that CryoGEM is capable of generating authentic cryo-EM images. The generated dataset can used as training data for particle picking and pose estimation models, eventually improving the reconstruction resolution.
\end{abstract}    

\vspace{-15pt}
\section{Introduction}
\label{sec:intro}

In the past decade, deep generative models like VAEs~\cite{vae}, GANs~\cite{gan}, and diffusion models~\cite{diffusion} have achieved significant success in conditional image generation. Recently, Stable Diffusion~\cite{stablediffusion} can produce high-quality images given simple textual descriptions. Additional controls \cite{controlnet} can be imposed to produce tailored visual effects, e.g., theatrical lighting~\cite{relighting} and specific perspectives~\cite{long2023wonder3d}. In fact, the successes of image generation have gone way beyond visual pleasantness, stimulating significant advances in scientific explorations. Examples include brain magnetic resonance imaging to computational tomography (MRI-to-CT) translation~\cite{mri2ct}, X-ray image generation~\cite{x-ray}, etc.  Different from entertainment applications, generation techniques for scientific imaging should faithfully follow the physical process: the generated results would eventually be applied to real-world downstream tasks such as medical image diagnosis. In this work, we extend image generation to a specific biomolecular imaging technique, single-particle cryo-electron microscopy (cryo-EM) to improve the performance of its downstream tasks. Cryo-EM aims to recover the near-atomic resolution 3D structure of proteins, with its latest application in recovering the SARS-COV-2 spike protein \cite{sars} structure for drug development. 

\begin{figure}[t]
    \centering
    \includegraphics[width=\linewidth]{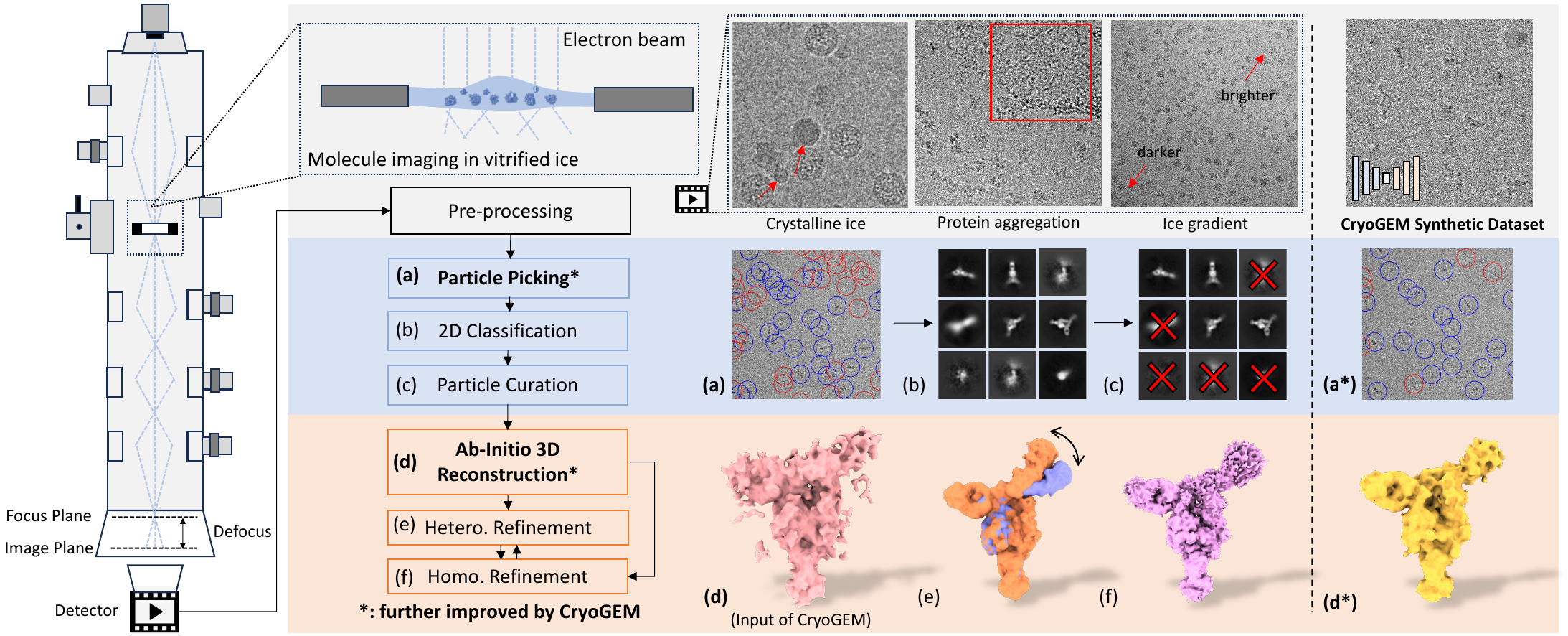}
    \caption{\textbf{CryoGEM improves cryo-EM data analysis.} Cryo-EM captures images of molecules in vitrified ice via electron beams. Data is processed for a high-resolution 3D reconstruction by a comprehensive pipeline. However, some modules like (a) particle picking and (d) ab-initio 3D reconstruction still lack high-quality training datasets. Given a coarse result as an input, \ours{} can synthesize authentic single-particle micrographs as training dataset augmentation.
    }
    \label{fig:teaser}
    \vspace{-15pt}
\end{figure}

As shown in Figure~\ref{fig:teaser}, a comprehensive data processing pipeline of single-particle cryo-EM starts with capturing transmission images of flash-frozen purified specimens, termed as \textit{micrographs}, using high-energy electron beams. The complete dataset contains hundreds of thousands of images of target particles with unknown locations, poses, and shapes. The main challenge in cryo-EM is to accurately estimate the locations and orientations of particles in the extremely low signal-to-noise ratio (SNR), mainly caused by detector shot noise in limited dosage conditions and other structural noises such as global ice gradients. Thus, the performance of existing methods for particle picking~\cite{topaz} and pose estimation~\cite{cryofire} are limited due to the lack of high-quality annotated training datasets, which are labor-intensive for human experts.

In this paper, we present a novel physics-informed generative cryo-electron microscopy (\ours{}) technique, which is capable of generating authentic annotated synthetic datasets using just 100 unannotated micrographs from a real dataset. To achieve highly controllable generation results, we introduced a simple yet highly controllable physical simulation process. Based on the coarse density volume, we achieve control at both the particle level and the micrograph level. However, this physical simulation still lacks critical authentic noise modeling. We thus adopt unpaired image-to-image translation to generate authentic noises. Existing methods like CycleGAN~\cite{cyclegan} assume that the source and target domains are bijective; however, real cryo-EM micrographs have random noises, existing contrastive learning methods, such as contrastive unpaired translation (CUT)~\cite{cut}, relax these assumptions by using a random sampling strategy for positive and negative samples. However, in cryo-EM, the randomly sampled positives and negatives can be semantically similar since particles are densely located in a micrograph. To address these challenges, we employ a novel contrast noise translation to transform the simulated micrographs into authentic cryo-EM micrographs. 

Furthermore, we use a particle-background segmentation paired with the simulated result as a guide for positive and negative sample selection. We show that precise guidance on sample selection can significantly improve the quality of image generation.

We validate \ours{} on five diverse and challenging real cryo-EM datasets. Extensive experiments show that \ours{} achieves significantly better visual quality compared with state-of-the-art methods. Also, we demonstrate that the performance of existing deep models in downstream tasks, including particle picking and particle pose estimation, can be significantly improved by training on our synthetic dataset. Notably, we achieve 44\% picking performance improvements, leading to 22\% better resolution of final reconstruction on average. Our code and datasets will be released.

\section{Related Works}
\label{sec:related_works}
Our work aims to extend generation methods to the field of cryo-EM. 
We therefore only discuss the most relevant works in respective fields. 

\noindent\textbf{Cryo-EM Pipeline.}
In recent decades, the CryoEM pipeline has rapidly evolved. Popular software such as CryoSPARC~\cite{cryosparc}, Relion~\cite{relion}, and Warp~\cite{warp} has been widely used for high-resolution 3D reconstruction of macromolecules. Two critical steps are particle picking and pose estimation. To achieve accurate particle picking, recent neural methods~\cite{topaz, cryotransformer, cryolo, cryoppp, cryosegnet, cryomae} have built upon several manually annotated real datasets or incorporated with few-shot learning techniques. However, they still lack generalization capability as the training dataset only covers a small portion of real scenarios. In pose estimation, traditional methods~\cite{relion, cryosparc} propose a Maximum-A-Posteriori (MAP) optimization by Expectation–Maximization (EM).
Recent self-supervised models~\cite{cryofire, cryoai, cryodrgn2} have been adopted for ab-initio reconstruction, i.e., 3D reconstruction from particle images with unknown poses. However, their performance is still limited without further fine-grained refinement processes. We improve the performance of particle picking and pose estimation by generating high-quality annotated synthetic datasets as training datasets.

\noindent\textbf{Cryo-EM Simulations.} Theoretical simulation techniques, based on physical priors, combine atomic-level simulations with global projection to accurately compute electron scattering during the imaging process \cite{insilicotem, Vulovic2013ImageFM, simtem, multem, abtem, Himes:rq5007}. Traditional simulation methods such as InsilicoTEM~\cite{insilicotem} model the interaction between electrons by taking specimens as multi-slices to improve the algorithm's performance. However, they typically require expensive computational resources and may require complex adjustments of parameters to achieve ideal results. Our work also includes an efficient physics-based simulation module to present the structural information of particles. Additionally, we generate realistic noises via a novel unpaired noise translation technique. 

\noindent\textbf{Unpaired Image-to-image Translation.} 
Deep generative models including GAN-based methods~\cite{cyclegan, dualgan, discogan, unit, munit} and diffusion models~\cite{unit-ddpm, cyclediff} are widely used in unpaired image-to-image translation tasks. CycleGAN~\cite{cyclegan} introduces generative adversarial networks to calculate cycle consistency losses, allowing for training on unpaired data~\cite{CycleGAN+VGG, DRITT++, LapMUNIT, UGAT-IT}. 
TraVeLGAN~\cite{travelgan}, DistanceGAN~\cite{distancegan}, and GcGAN~\cite{gcgan} achieve one-way translation while avoiding traditional cycle consistency.
Diffusion models~\cite{choi2021ilvr, unit-ddpm, meng2021sdedit, cyclediff, zhao2022egsde, cycle-net, diff-ae} have been recently introduced to unpaired image-to-image translation. But they often demonstrate results in low resolution and they often rely on a large-scale training dataset. Recently, Contrastive Unpaired Translation (CUT)~\cite{cut} introduces a new generative framework via contrastive learning~\cite{SRC, NEGCUT, FeSim} to propose a more efficient training framework. However, existing methods do not combine the physical process as additional constraints.
In contrast, our method includes a physical simulation during authentic cryo-EM micrograph generation.

\section{Physics-informed Generative Cryo-EM}

\begin{figure*}[t]
  \centering
   \includegraphics[width=\linewidth]{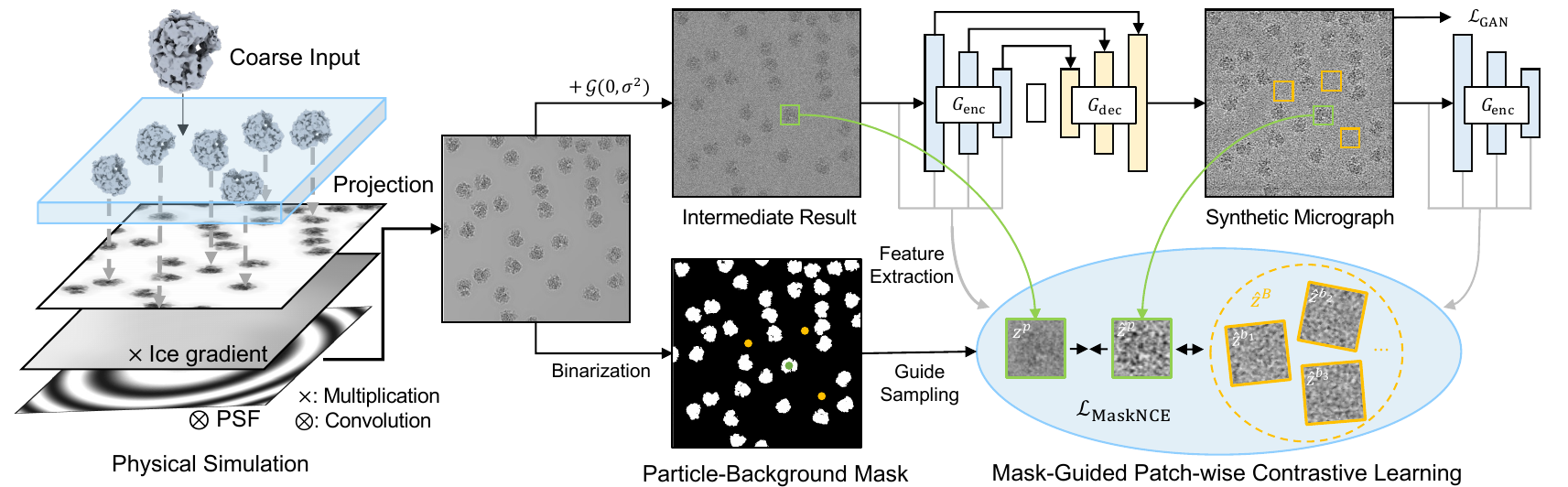}
   \caption{\textbf{Pipeline of \ours{}.}
  We begin by creating a virtual specimen containing various initial reconstruction results. We then simulate the imaging process of cryo-EM, incorporating physical priors such as ice gradient and point spread function (PSF) to generate a physical simulation. By adding simple Gaussian noise to the physically simulated results, we introduce randomness within a contrastive learning framework. To enhance training efficiency and performance, we use the particle-background mask as a guide for patch sampling. The sampled positive and negative instances are then encoded into multi-scale features for contrastive learning. Additionally, we introduce an adversarial loss to ensure realistic cryo-EM image synthesis.
   }
   \label{fig:pipeline}
\end{figure*}

We propose \ours{}, the first method to combine a physics-based simulation process with a novel contrastive noise generation technique (Figure~\ref{fig:pipeline}). Our approach begins with preparing a virtual specimen containing numerous uniformly distributed target protein structures (Section~\ref{sec:method:vitural-specimen-construction}). We then emulate the cryo-EM imaging process to introduce physical constraints (Section~\ref{sec:imaging-process}). To generate authentic cryo-EM noise, we use a novel unpaired noise translation technique via contrastive learning guided by a particle-background mask and detail the final objective during training. (Section~\ref{sec:contrastive-noise-generation})

\subsection{Virtual Specimen Preparation}
\label{sec:method:vitural-specimen-construction}

To simulate the cryo-EM imaging process, we first prepare the virtual specimen $S(x, y, z)$, a large 3D density volume that contains multiple copies of the target molecule's coarse result with randomly generated locations, orientations, and conformations. To obtain the coarse result, we employ cryoSPARC \cite{cryosparc} for ab-initio reconstruction, followed by cryoDRGN \cite{cryodrgn} for a continuous heterogeneous reconstruction to obtain a neural volume
\begin{equation}
    V(\mathbf{p}, \mathbf{w}): \mathbb{R}^3 \times \mathbb{R}^8 \mapsto \mathbb{R},
\end{equation}
where $\mathbf{p} = (x,y,z)^{\text{T}} \in \mathbb{R}^3$ represents the spatial coordinates, $\mathbf{w}\in \mathbb{R}^8$ denotes a high-dimensional conformational embedding. This neural volume can generate a density volume given a learned conformational embedding following CryoDRGN~\cite{cryodrgn}. Notably, such a coarse result can be easily resolved, but the final result requires further iterative optimizations of selected particles, estimated particle poses, 3D templates, etc.

During the specimen preparation, we sequentially add $N$ particles $V(\mathbf{p}, \mathbf{w})$ into an empty virtual specimen. During every addition, we randomly sample orientation matrices $R_i\in\mathbb{R}^{3\times 3}$ from SO(3) space, random conformations $\mathbf{w}_i$ from the learned conformational space, as well as translations $\mathbf{t}_i\in\mathbb{R}^{3}$ from unoccupied areas of the virtual specimen. Both sampled orientations and locations can be further used as ground-truth annotations. Thus, the virtual specimen can be expressed as:
\begin{equation}
    S(x, y, z) = \sum_{i=1}^{N}V(R_i\mathbf{p}+\mathbf{t}_i; \mathbf{w}_i),
    \label{equ:virual-specimen}
\end{equation}
where the number of particles \(N \sim \mathcal{N}(\mu_N, \sigma_N^2)\), with the mean \(\mu_N\) and standard deviation \(\sigma_N\) derived from the actual distribution of particle count in real micrographs. We ensure that the minimum \(N_{\min}\) and maximum \(N_{\max}\) values are within two standard deviations from the mean.

\subsection{Emulating the Imaging Process}
\label{sec:imaging-process}

We emulate the physical imaging process of cryo-EM including electron-specimen interaction, ice gradient, and cryo-EM optical aberration based on virtual specimen. For simplicity, we consider the complex electron-specimen interaction as an orthogonal projection of the specimen by applying the weak phase object approximation~\cite{Williams1996TransmissionEM} (WPOA). To obtain the true ice gradients, we estimate them on real micrographs by IceBreaker~\cite{olek2022icebreaker}, which can generate a weight map of ice, $W(x,y)$, whose every pixel represents the attenuation ratio compared to the maximum value of intensity. Also, we model the optical aberrations as a point spread function (PSF) $g$, which is the Fourier transform of the contrast transfer function (CTF), by off-the-shelf software CTFFIND4~\cite{rohou2015ctffind4}.
To sum up, the 2D physics-based result $I_{\text{phy}}(x,y)$ can be expressed as:

\begin{equation}
    I_{\text{phy}}(x, y) = g * \left[W(x,y)\cdot\int_{\mathbb{R}}S(x, y, z)\mathrm{d}z\right].
\end{equation}

During the simulation process, we randomly select a pair of weight maps and PSF estimated from real micrographs.

\begin{figure*}[t]
  \centering
  \includegraphics[width=\linewidth]{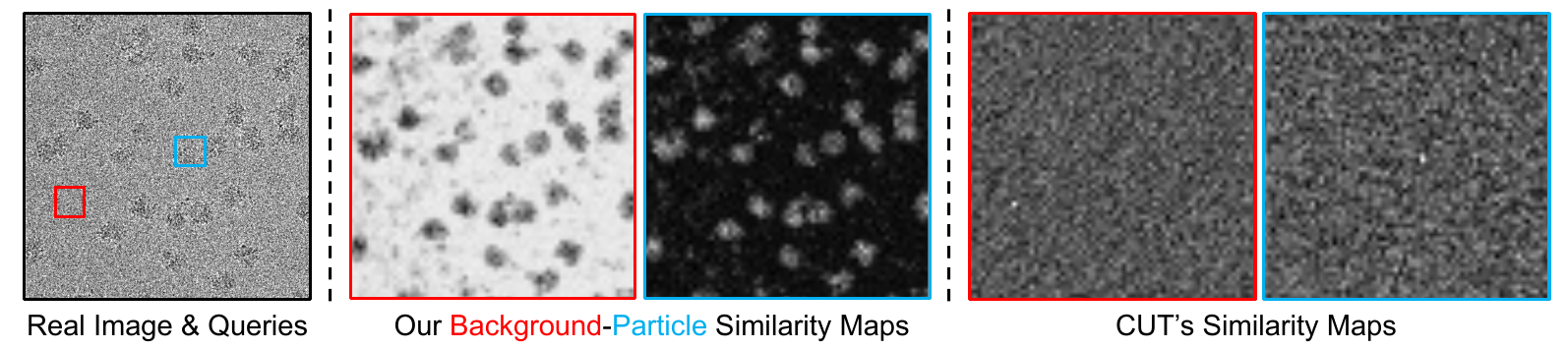}
   \caption{\textbf{Visualization of the learned similarity.} Given query patches (red for particle and blue for background) on the input real micrograph, we visualize the learned similarity maps of our and CUT's encoders $G_{enc}$ by calculating exp($G_{enc}(v)\cdot G_{enc}(v^{-}) / \tau$), where $v$ denotes the query and $v^{-}$ denotes the patches of real micrograph. The results imply that our encoder can recognize particles and backgrounds in real cryo-EM micrographs. However, CUT fails to learn that without the guidance of particle-background maps during training.
   }
   \label{fig:similarity}
\end{figure*}

\section{Contrastive Noise Generation}
\label{sec:contrastive-noise-generation}
Real cryo-EM micrographs contain high-level noises related to specimen-electron interaction and detector~\cite{Vulovic2013ImageFM}. Thus, accurate noise modeling is essential for authentic synthetic cryo-EM micrograph generation.
We model the noise generation as an unpaired image-to-image translation task. The input, $I_{\text{phy}} \sim X$, learns the actual noise distribution from real cryo-EM images $I_{\text{real}} \sim Y$, to generate synthetic cryo-EM images $I_{\text{syn}} \sim \hat Y$. The generative model includes a discriminator \( D \), and a generator \( G = G_{\text{dec}} \circ G_{\text{enc}} \), where \( G_{\text{enc}} \) is a encoder and \( G_{\text{dec}} \) is a decoder. 

\noindent\textbf{Non-deterministic noise translation.} Existing contrastive learning methods such as CUT~\cite{cut} have shown significant promise for efficient training and generating high-quality natural images. Nonetheless, they do not account for generating random noise in cryo-EM, i.e., they can only generate synthetic micrographs in a deterministic way, against the nature of the random noise generation process, leading to an unstable training process and degraded performance. To address this, we introduce a random process by adding a zero-mean Gaussian random noise $\mathcal{G}$ into the physical simulations. Thus, the synthetic image can be defined as:
\begin{equation}
I_{\text{syn}} = G(I_{\text{phy}} + \mathcal{G})
\end{equation} 
This simple strategy significantly improves our noise generation quality by constructing a mapping of a single physical simulation result to infinite synthetic noisy results with varied noise patterns. We also show that the content is still well preserved after the translation as they provide the mutual information between source and target.

\noindent\textbf{Mask-Guided sampling scheme.} A fundamental assumption of patch-wise contrastive learning is that the randomly chosen negatively sampled patches are semantically different from the positive samples~\cite{debiasedcontrastivelearning}. But in a single cryo-EM micrograph, hundreds of particles are densely distributed against an almost pure yet noisy background, which means that randomly chosen negative and positive patches are likely semantically similar. Taking advantage of our highly controllable physical simulation process, we introduce a mask-guided sampling scheme by generating a binary particle-background mask paired with $I_{phy}$, denoted as 
$M(x,y)$, where $M(x,y)=1$ indicates a particle. Guided by this mask, our selection process can choose positive and negative samples from locations with contrasting mask labels. This significantly improves the encoder's performance, as shown in Figure~\ref{fig:similarity}, our encoder can generate accurate similarity maps given particles or backgrounds as references on real images, but CUT's encoder fails to recognize them.

\noindent\textbf{Mutual information extraction.} Guided by particle-background mask, we select $Q$ paired patches in \(I_{\text{phy}}\) and \(I_{\text{syn}}\) as the positive samples and queries on particle region, respectively. We then choose \(K\) patches in \(I_{\text{phy}}\) as negative samples on the background region. The encoder \(G_{\text{enc}}\) maps these to normalized vectors \(\boldsymbol{v} = \{\boldsymbol{v}_q\}_{q=1}^{Q}\), \(\boldsymbol{v}^+ = \{\boldsymbol{v}^{+}_q\}_{q=1}^{Q}\), and $\boldsymbol{v}^{-} = \left\{\boldsymbol{v}_k^{-}\right\}_{k=1}^{K}$, respectively. Our objective is to maximize the likelihood of selecting a positive sample by minimizing the cross-entropy loss:
\begin{equation}
\ell(\boldsymbol{v}, \boldsymbol{v}^+, \boldsymbol{v}^-) = 
-\sum_{q=1}^{Q}\log \left[ \frac{\exp(\boldsymbol{v}_q \cdot \boldsymbol{v}^+_q / \tau)}{\exp(\boldsymbol{v}_q  \cdot \boldsymbol{v}_q ^+ / \tau) + \sum_{k=1}^{K} \exp(\boldsymbol{v}_q  \cdot \boldsymbol{v} _k^- / \tau)} \right],
\end{equation}
where \(\tau\) is a temperature factor that scales the distance between the query and samples.

\textbf{Mask-guided patch-wise contrastive learning.} We leverages multiple intermediate layers of \( G_{\text{enc}} \) to extract multi-scale features from input patches. Notably, we align the receptive field with particle size by selecting the \( G_{\text{enc}} \)'s first \(L\) layers. Feature map from each layer is then passed through compact two-layer MLP networks \( H_l \) (\( l \in \{1,2,..., L\} \)) to generate a feature \(\boldsymbol{z}_l=H_l(G_{\text{enc}}^l(I_{\text{inter}}))\), where \( G_{\text{enc}}^l \) is the \( l \)-th layer. We define those on particles as \( p \in P \), and those on the background as \( b \in B \). The feature of \( l \)-th layer at a particle position \( p \) is denoted as \( \boldsymbol{z}^{p}_l \), and the set represented as \( \boldsymbol{z}^{P}_l = \left\{\boldsymbol{z}^{p}_l\right\}_{p \in P} \). Similarly, for background positions, we denote the \( l \)-th layer feature as \( \boldsymbol{z}^{b}_l \), with the set represented as \( \boldsymbol{z}^{B}_l = \left\{\boldsymbol{z}^{b}_l\right\}_{b \in B} \). Features encoded from the output \( I_{\text{syn}} \) are represented as \( \hat{\boldsymbol{z}} \), where \(\hat{\boldsymbol{z}}_l=H_l(G_{\text{enc}}^l(I_{\text{syn}}))\). We propose a novel mask-guided Noise Contrastive Estimation (NCE) loss for efficient contrastive learning in cryo-EM:
\begin{equation} 
     \mathcal{L}_{\text{MaskNCE}}(G, H, X|\mathcal{G}) = 
     \mathbb{E}_{I_{\text{inter}}\sim X|\mathcal{G}} 
     \sum_{l=1}^L\Biggl[\sum_{p \in P}\ell\left(\boldsymbol{z}_l^p, \hat{\boldsymbol{z}}^p_l, \hat{\boldsymbol{z}}^B_l\right)
     +\sum_{b \in B}\ell\left(\boldsymbol{z}_l^b, \hat{\boldsymbol{z}}^b_l, \hat{\boldsymbol{z}}^P_l\right)\Biggr].
\label{equ:masknce}
\end{equation}

By minimizing $\mathcal{L}_{\text{MaskNCE}}$, we effectively differentiate particle and background features (Figure~\ref{fig:similarity}).

\noindent\textbf{Final objective.} We introduce an adversarial loss function, \(\mathcal{L}_{\text{GAN}}\), to encourage the network to produce more realistic simulated cryo-EM images, \(I_{\text{syn}}\), as follows:

\begin{equation}
\mathcal{L}_{\text{GAN}}(G, D, X|\mathcal{G}, Y) =  \mathbb{E}_{I_{\text{real}} \sim Y}[\log D(I_{\text{real}})] \nonumber  + \mathbb{E}_{I_{\text{inter}} \sim X|\mathcal{G}}[\log (1 - D(G(I_{\text{inter}})))].
\end{equation}

Combining this with the contrastive learning loss \(\mathcal{L}_{\text{MaskNCE}}\), the overall training loss function is:
\begin{equation}
    \mathcal{L}_{\text{GAN}}(G,D,X|\mathcal{G},Y)+\lambda\mathcal{L}_{\text{MaskNCE}}(G,H,X|\mathcal{G}),
\end{equation}
we set the hyper-parameter $\lambda$ to $10.0$ in all our experiments.

\definecolor{best1}{RGB}{222,242,212}
\definecolor{best2}{RGB}{255,250,212}

\begin{figure}[t]
  \centering
  \includegraphics[width=\linewidth]{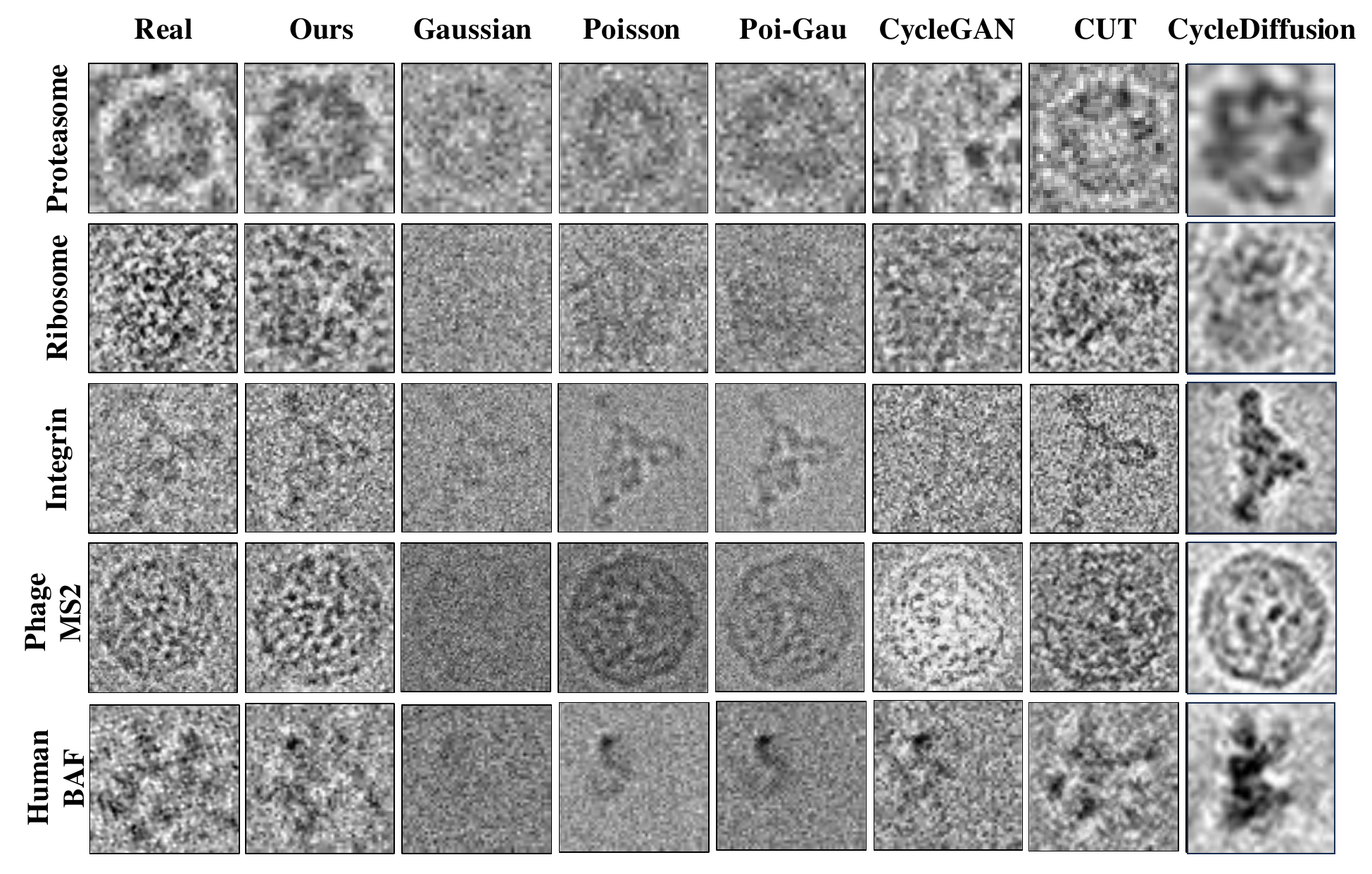}
   \caption{\textbf{Qualitative comparison results.} 
   Our approach achieves the most authentic noise generation across all datasets. The traditional noise models succeed in preserving the structural information while lacking realistic noise patterns. CycleGAN, CUT, and CycleDiffusion introduce severe artifacts on generated results.}
   \label{fig:results}
\end{figure}

\section{Experiments}\label{sec:exp}
\paragraph{Datasets.}
We evaluate \ours{} across five challenging cryo-EM datasets. Each includes expert-curated particle-picking annotations for high-resolution reconstruction of target molecules. Note that the particle annotations in the dataset are solely used for validation of downstream tasks and are not utilized during the training phase of our \ours{} model. 1) T20S \textbf{Proteasome} (EMPIAR-10025)~\cite{Campbell201528R}' micrographs exhibit a high density of particles, leading to occlusions. posing significant challenges in particle picking. Also, the 3D structure is D7 symmetric, which brings ambiguities in the pose estimation task. 2) 80S \textbf{Ribosome} (EMPIAR-10028)~\cite{Wong2014CryoEMSO} has a complex 3D structure, making it difficult to estimate the poses of particle images. 3) Asymmetric \(\alpha\text{V}\beta 8\) \textbf{Integrin} (EMPIAR-10345)~\cite{Campbell2020CryoEMRI} contains molecules with varied conformations, requiring a heterogeneous reconstruction. 4) \textbf{PhageMS2} (EMPIAR-10075)~\cite{empiar10075, cryoppp}'s micrographs contain enormous spherically-shaped virus particles, suitable for testing the generalizability of the particle picking model. 5) Endogenous \textbf{HumanBAF} complex (EMPIAR-10590)~\cite{humanbaf, cryoppp} has significant ice gradient artifacts. We kindly refer to Appendix~\ref{app:data-pre} for a more detailed dataset description and illustration.

\noindent\textbf{Baselines.} 
We evaluate the performance of \ours{} compared to several traditional noise baselines and deep generative models. We use \textbf{Poisson} noise to simulate the high-level detector shot noises in cryo-EM images, \textbf{Gaussian} noise for a general simplified noise modeling in cryo-EM, and Poisson-Gaussian mixed noise (\textbf{Pos-Gau}) as traditional baselines. We choose \textbf{CycleGAN}, \textbf{CUT}, and recent \textbf{CycleDiffusion} as deep generative baselines. In Appendix~\ref{app:baseline}, we detail their specific settings.

\noindent\textbf{Implementation details.} All experiments are conducted on a single NVIDIA GeForce RTX 3090 GPU. As a lightweight model, \ours{} trains 100 epochs on a single dataset within an hour using less than 10 GB of memory. The training dataset (all held out for the evaluation) is comprised of 100 real images alongside 100 physics-based simulated results. We enhance the dataset through data augmentation, specifically by rotating the images by 90, 180, and 270 degrees. Please see Appendix B for more details. We choose PatchGAN \cite{isola2017image} as our discriminator and UNet implemented by~\cite{cut} as our generator. We fix the image resolution to $1024 \times 1024$ during training. Guided by the particle-background mask, we evenly sample 256 queries on particles and 256 on the background, ensuring that their corresponding negative samples are located where the labels are opposite.

\begin{figure}[t]

  \centering
  \includegraphics[width=\linewidth]{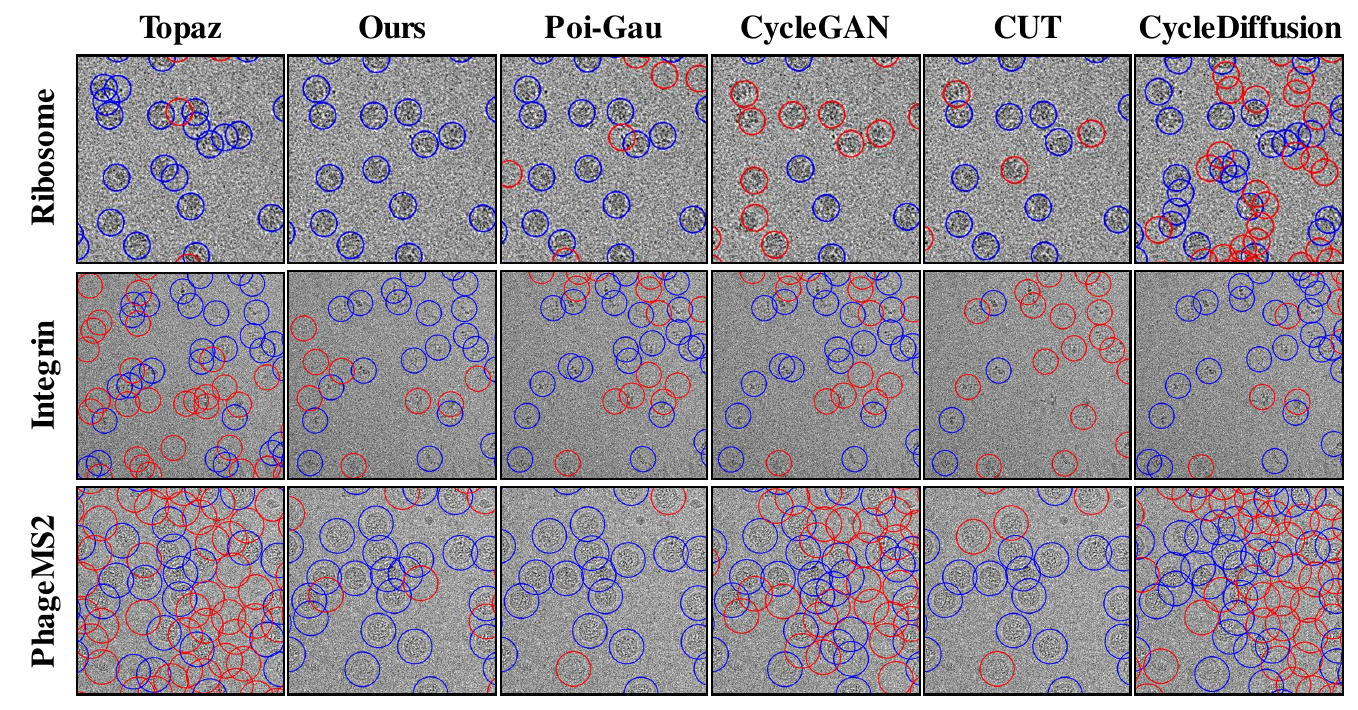}
   \caption{\textbf{Qualitative comparison results of particle picking.}
   The blue circles indicate matches with manual picking results, while the red circles represent misses or excess picks by the model.
   }
   \label{fig-sup:particle_picking}
\end{figure}

\begin{table}[t]
\caption{\textbf{Quantitative comparison of visual quality.} 
Our approach consistently achieves the best performance in FID metric.}
\label{tab:fid}
\centering
\resizebox{!}{0.45in}{%
\begin{tabular}{ccccccc}
\hline
Metric          & \multicolumn{5}{c}{FID$\downarrow$} \\ \hline
Dataset         & Proteasome & Ribosome & Integrin & PhageMS2 & HumanBAF & Avg.        \\ \hline
Gaussian        & 89.87  & 177.01 & 174.94 & 162.81 & 46.48  & 130.22 \\
Poisson         & 112.40 & 86.88  & 173.94 & 343.16 & 44.17  & 152.11 \\
Poi-Gau         & 93.41  & 73.89  & 173.64 & 311.50 & 45.55  & 139.60 \\
CycleGAN        & 89.33  & 27.73  & 54.83  & 422.80 & 137.41 & 146.42 \\
CUT             & 46.61  & 44.23  & 49.96  & 88.65  & 74.76  & 60.84  \\ 
CycleDiffusion  & 470.37 & 173.62 & 386.83 & 468.83 & 577.46 & 415.42 \\ \hline
Ours            & \textbf{42.96}  & \textbf{6.54}  & \textbf{42.46} & \textbf{63.11}  & \textbf{34.50}  & \textbf{37.91}   \\ \hline
\end{tabular}
}
\end{table}
\begin{table}[t]
\caption{\textbf{Quantitative comparison of particle picking.} 
Our approach consistently achieves the best in AUPRC and Res(\AA) metrics.}%
\label{tab:partical_picking}
\centering
\resizebox{!}{0.44in}{%
\begin{tabular}{ccccccccccccc}
\hline
Metric & \multicolumn{6}{c}{AUPRC$\uparrow$} & \multicolumn{6}{c}{Res(\AA) $\downarrow$} \\ \hline
Dataset  & Proteasome & Ribosome & Integrin & PhageMS2 & HumanBAF & Avg. & Proteasome & Ribosome & Integrin & PhageMS2 & HumanBAF & Avg. \\ \hline
Gaussian   & 0.471 & 0.485 & 0.259 & 0.886 & 0.470 & 0.514 & 2.76 & 3.84 & 7.62 & 7.44 & 10.29 & 6.39 \\
Poisson    & 0.469 & 0.375 & 0.213 & 0.490 & 0.483 & 0.406 & 2.77 & 4.16 & 7.52 & 10.50& 10.67 & 7.12 \\
Poi-Gau    & 0.455 & 0.618 & 0.210 & 0.706 & 0.381 & 0.474 & 2.80 & 3.87 & 8.13 & 10.90& 13.18 & 7.77 \\
CycleGAN   & 0.200 & 0.308 & 0.414 & 0.895 & 0.228 & 0.409 & 5.10 & 3.93 & 8.03 & 7.51 & 11.83 & 7.28 \\
CUT        & 0.442 & 0.224 & 0.335 & 0.592 & 0.513 & 0.421 & 2.77 & 4.78 & 7.16 & 8.91 & 12.92 & 7.30 \\
CycleDiffusion 
           & 0.346 & 0.205 & 0.233 & 0.469 & 0.392 & 0.329 & 2.92 & 4.56 & 6.35 & 9.51 & 6.68 
& 6.66 \\ \hline
Topaz      & 0.302 & 0.679 & 0.526 & 0.329 & 0.493 & 0.466 & 3.13 & 3.84 & 6.34 & 10.67 & 9.59 & 6.71 \\ \hline
Ours       & \textbf{0.490} & \textbf{0.797} & \textbf{0.606} & \textbf{0.915} & \textbf{0.562} & \textbf{0.674} & \textbf{2.68} & \textbf{3.25} & \textbf{5.54} & \textbf{7.16} & \textbf{7.74} & \textbf{5.27}   \\ \hline
\end{tabular}%
}
\end{table}
\begin{table}[t]
\caption{\textbf{Quantitative comparison of pose estimation.} Our approach achieves the best performance in Res(px) and Rot.(rad) metrics.}
\label{tab:pose}
\centering
\resizebox{!}{0.47in}{%
\begin{tabular}{ccccccccccccc}
\hline
Metric   & \multicolumn{6}{c}{Res(px)$\downarrow$} & \multicolumn{6}{c}{Rot.(rad)$\downarrow$} \\ \hline
Dataset  & Proteasome & Ribosome & Integrin & PhageMS2 & HumanBAF & Avg. & Proteasome & Ribosome & Integrin & PhageMS2 & HumanBAF & Avg. \\ \hline
Gaussian & 2.97 & 4.59  & 7.01  & 5.85  & 6.82 & 5.44 & 0.48 & 0.50 & 1.20 & 0.64 & 1.49 & 0.88 \\
Poisson  & 3.02 & 4.912 & 7.01  & 5.98  & 8.03 & 5.79 & 1.20 & 0.90 & 1.19 & 0.69 & 1.47 & 1.10 \\
Poi-Gau  & 3.02 & 4.39  & 9.06  & 5.90  & 8.12 & 6.09 & 1.05 & 0.39 & 1.40 & 0.61 & 1.43 & 0.98 \\
CycleGAN & 3.02 & 4.74  & 6.24  & 5.71  & 6.91 & 5.32 & 0.46 & 0.61 & 1.55 & 0.74 & 1.48 & 0.97 \\
CUT      & 2.66 & 5.40  & 6.13  & 6.03  & 9.58 & 5.96 & 0.44 & 1.15 & 1.53 & 0.66 & 1.45 & 1.05 \\ 
CycleDiffusion 
         & 3.61 & 5.79  & 8.5   & 5.91  & 9.33 & 6.62 & 0.44 & 1.42 & 1.55 & 0.58 & 1.53 & 1.10 \\ \hline
CryoFIRE & 5.94 & 16.92 & 13.87 & 17.23 & 6.98 & 12.18& 1.55 & 0.64 & 0.93 & 0.75 & 1.53 & 1.08 \\ \hline
Ours     & \textbf{2.59} & \textbf{4.27}  & \textbf{4.88}  & \textbf{5.54}  & \textbf{6.56} & \textbf{4.29} & \textbf{0.41} & \textbf{0.32} & \textbf{0.88} & \textbf{0.43} & \textbf{1.42} & \textbf{0.69}  \\ \hline
\end{tabular}%
}
\end{table}

\subsection{Visual Quality}
We evaluate \ours{}'s visual quality compared to baselines. As illustrated in Figure~\ref{fig:results}, our approach consistently achieves the best qualitative results across all datasets in terms of overall noise patterns and the preservation of particle structural information. Traditional methods fail to mimic authentic noise distributions. CycleGAN struggles with convergence due to the breakdown of its bijection assumption. CUT can achieve better visual quality but its performance is still limited due to its random sampling scheme. CycleDiffusion cannot learn the high-frequency structural information and noise patterns.
As shown in Table~\ref{tab:fid}, we also employed Frechet Inception Distance (\textbf{FID}) which is widely used in generative tasks to measure the similarity between generated and real data. Our results significantly outperform existing baseline methods in all five datasets. For additional results, please refer to Figure~\ref{fig-sup:results} and Figure~\ref{fig-sup:gallery} in the appendix.

\subsection{Particle Picking}
To evaluate how \ours{} can enhance downstream particle picking task, we finetune the popular particle picking model \textbf{Topaz} \cite{topaz} using synthetic annotated datasets from different baselines. Our approach outperforms the original Topaz and other baseline methods in terms of the quality of visual picking results. As illustrated in Figure~\ref{fig-sup:particle_picking}, the fine-tuned Topaz model is employed to select particles across five datasets. The original Topaz often mistakenly picks false particles from contaminants, particle aggregations, and ice patches. Our fine-tuned Topaz significantly improves accuracy by reducing the incidence of false positives and redundant particle picks. Table \ref{tab:partical_picking} presents the quantitative particle picking results, where we utilize the Area Under the Precision-Recall Curve (\textbf{AUPRC}) metric that is widely used as particle picking metric~\cite{topaz,cryolo}. For the calculation of the AUPRC score, we retain the original particle-picking results for comparison against the ground truth labels, without any thresholding.  The quantitative comparison shows that the finetuned Topaz using our data achieves the best performance across all datasets. 
We also show that our final reconstruction resolution (\textbf{Res(\AA)}) is significantly improved using cryoSPARC~\cite{cryosparc}'s default reconstruction pipeline.
Notably, we present the coarse volumes used for \ours{} training, the final resolved structures from \ours{} particle picking workflow, and the FSC curves from different baselines in Figure~\ref{fig-sup:pp-volumes} in the Appendix~\ref{app:pp-details}.

\subsection{Pose Estimation}
In the cryo-EM reconstruction pipeline, the accuracy of pose estimation is crucial for the resolution of the final reconstruction. Existing cryo-EM ab-initio methods, such as CryoFIRE~\cite{cryofire}, leverage a self-supervised learning framework to predict particle poses and reconstruct 3D volume from images simultaneously. Given synthetic datasets generated from different baselines, we directly supervise its pose estimation module by minimizing the quadratic error between the matrix entries in the prediction and ground truth. The details of direct supervision and the evaluation metrics as well as the visualization of improved structures (Figure~\ref{fig-sup:pe-improve}) can be found in Appendix~\ref{app:pe-details}.
We evaluate the pose estimation module on real datasets to obtain estimated poses as ground truths. We use a traditional reconstruction algorithm, filter back-projection (FBP), to obtain final reconstruction results. 
As demonstrated in Table~\ref{tab:pose}, the pre-trained pose estimation module using our data consistently achieves the best performance in terms of pose accuracy (rotation error in radian) and reconstruction resolution (in pixel). 

\begin{figure}[t]
    \vspace{-10pt}
    \centering
    \begin{minipage}[t]{0.5\textwidth}
        \centering
        \raisebox{1\height}{\resizebox{\textwidth}{!}{%
            \begin{tabular}{cccccc}
            \\ \hline
            Dataset              & \multicolumn{5}{c}{Ribosome}       \\ \hline
            Task &
              \multicolumn{1}{c}{Photo-realism} &
              \multicolumn{2}{c}{Particle Picking} &
              \multicolumn{2}{c}{Pose Estimation} \\ \hline
            Metric &
              \multicolumn{1}{c}{FID$\downarrow$} &
              \multicolumn{1}{c}{AUPRC$\uparrow$} &
              \multicolumn{1}{c}{Res(\AA)$\downarrow$} &
              \multicolumn{1}{c}{Res(px)$\downarrow$} &
              Rot(rad)$\downarrow$ \\ \hline
            w/o Physical Priors  & 16.63 & \underline{0.789} & 3.29 & 4.88 & 0.79 \\ \hline
            Third Layer Noise    & 15.83 & 0.764 & 3.31 & 4.45 & 0.47 \\
            Fifth Layer Noise    & 21.30 & 0.777 & \underline{3.26} & 4.46 & 0.51 \\
            Second Channel Noise & \underline{6.63}  & 0.012 & 9.76 & 4.29 & 0.34 \\ \hline
            SNR=0.01             & 22.16 & 0.313 & 3.80 & 7.11 & 1.50 \\
            SNR=1.0              & 17.95 & 0.749 & 3.40 & \textbf{4.11} & \textbf{0.25} \\
            w/o Gaussian Noise   & 50.74 & 0.735 & 3.31 & 4.27 & 0.35 \\ \hline
            w/o Particle         & 6.85  & 0.788 & 3.41 & 4.65 & 0.50 \\
            w/o Background       & 18.46 & 0.779 & 3.39 & 4.33 & 0.34 \\
            w/o Mask Guide       & 12.61 & 0.721 & 3.29 & 4.44 & 0.41 \\ \hline
            Ours                 & \textbf{6.54}  & \textbf{0.797} & \textbf{3.25} & \underline{4.27} & \underline{0.32} \\ \hline
            \end{tabular}
        }}
    \end{minipage}
    \hfill
    \begin{minipage}[t]{0.38\textwidth}
        \centering
        \hspace{-1.5cm}
        \raisebox{-0.2cm}{
        \includegraphics[width=\linewidth]{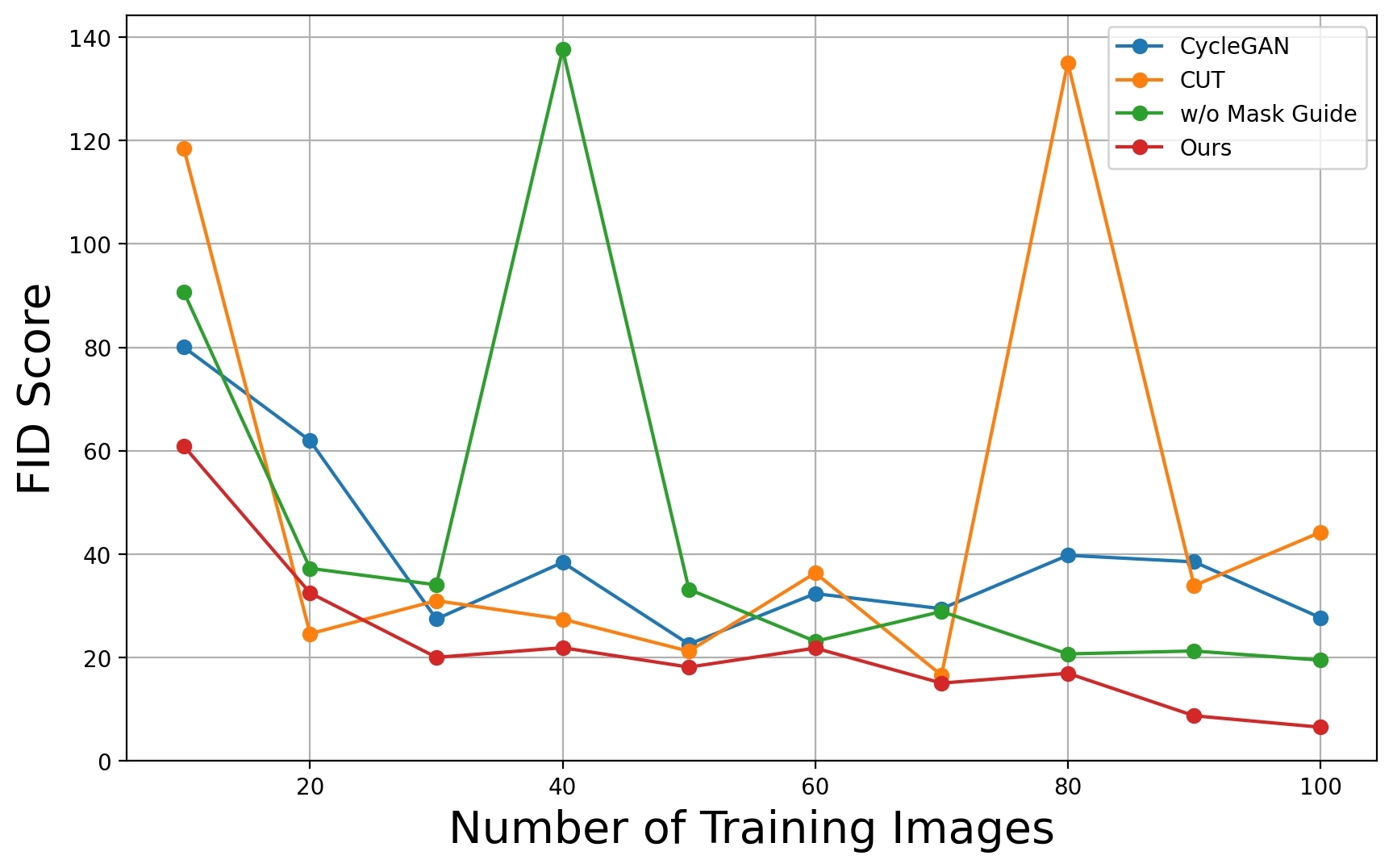}
        }
    \end{minipage}
    \caption{\textbf{Quantitative evaluation of ablation study.} Our complete model consistently achieves the \textbf{best} or \underline{second} performance across all metrics, as shown in the left table. It demonstrates stable and efficient training compared to CycleGAN, CUT, and the ablation without Mask Guide (w/o Mask Guide), as illustrated in the right figure.}
    \label{fig:ablation}
    \vspace{-10pt}
\end{figure}

\subsection{Ablation Study}
We validate the effectiveness of our several key designs on the Ribosome dataset.

\noindent\textbf{Physical priors.}
We denote the variation of our model without physical priors as \textbf{w/o Physical Priors}. Results imply that the physical priors narrow the domain gap between simulated results and real micrographs.

\noindent\textbf{Introducing randomness.}
We introduce randomness by different kinds of noises, such as introducing feature-level noise in the encoder's third layer (\textbf{Third Layer Noise}) and fifth layer (\textbf{Fifth Layer Noise}). Additionally, we treat random noise as a second input channel, which may potentially preserve the content of physical simulation better (\textbf{Second Channel Noise}). Also, we demonstrate the impact of noise scaling relative to the signal with \textbf{SNR=0.01} and \textbf{SNR=1.0}. Additionally, we remove the Gaussian noise (\textbf{w/o Gaussian Noise}) to validate the effectiveness of this randomness.

\noindent\textbf{Sampling scheme.}
We use a particle-background mask to guide sampling in the contrastive learning process. For accurate particle position and shape control, the generator must preserve the input image's content. If we sample positives only in the background, the mutual information between particles can only be optimized indirectly, leading to inaccurate particle positions and shapes (\textbf{w/o Particle}). Conversely, sampling positives only in particles results in less realistic noise patterns (\textbf{w/o Background}). We also present the results without the mask-guided sampling scheme (\textbf{w/o Mask Guide}), which includes only the physics-based module and input domain randomness.

\noindent\textbf{Training efficiency ablation.}
In the right of Figure~\ref{fig:ablation}, we validate that our method can generate realistic cryo-EM images with fewer samples compared to baselines in terms of FID. Notably, some high FID outliers indicate that removing the mask-guided sampling scheme or ignoring the physics-based module can decrease training stability.
\section{Conclusion}
\label{sec:conclusion}

\noindent\textbf{Limitations.} As the first trial to achieve authentic cryo-EM image generation with a novel combination of physics-based simulation and unpaired noise translation via contrastive learning, \ours{} does have room for further improvements. First, our physics-based simulation relies on a coarse result as an input, which may be hard to obtain in some challenging cases, e.g., when the target molecule is very small or dynamic. The latest AlphaFold 3~\cite{abramson2024accurate} can be helpful to directly predict the structure. Furthermore, we sacrifice the generalization capability for a lightweight training framework. In the future, we will train a generalized version of \ours{} to unlock more real applications.

\noindent\textbf{Conclusion.}
We have presented the first generative approach in cryo-EM image synthesis, \ours{}, that marries physics-based simulation with a contrastive noise generation, to enhance downstream deep models, finally improving cryo-EM reconstruction results.  Extensive experiments have shown that \ours{}'s generated dataset with ground-truth annotations can effectively improve particle picking and pose estimation models, eventually improving reconstruction results. We believe that \ours{} serves as a critical step for high-fidelity cryo-EM data generation using deep generative models, with diverse applications of structure discovery in in cryo-EM.

{
    \small
    \bibliographystyle{plain}
    \bibliography{main}
}
\clearpage
\begin{center}
\LARGE\textbf{---Supplementary Material---\\
\ours{}: Physics-Informed Generative Cryo-Electron Microscopy}
\end{center}
\vspace{9pt}
\setcounter{section}{0}
\renewcommand{\thesection}{\Alph{section}}

\section{Additional results.}
\begin{figure}[htbp] 
    \centering
    \includegraphics[width=0.95\textwidth]{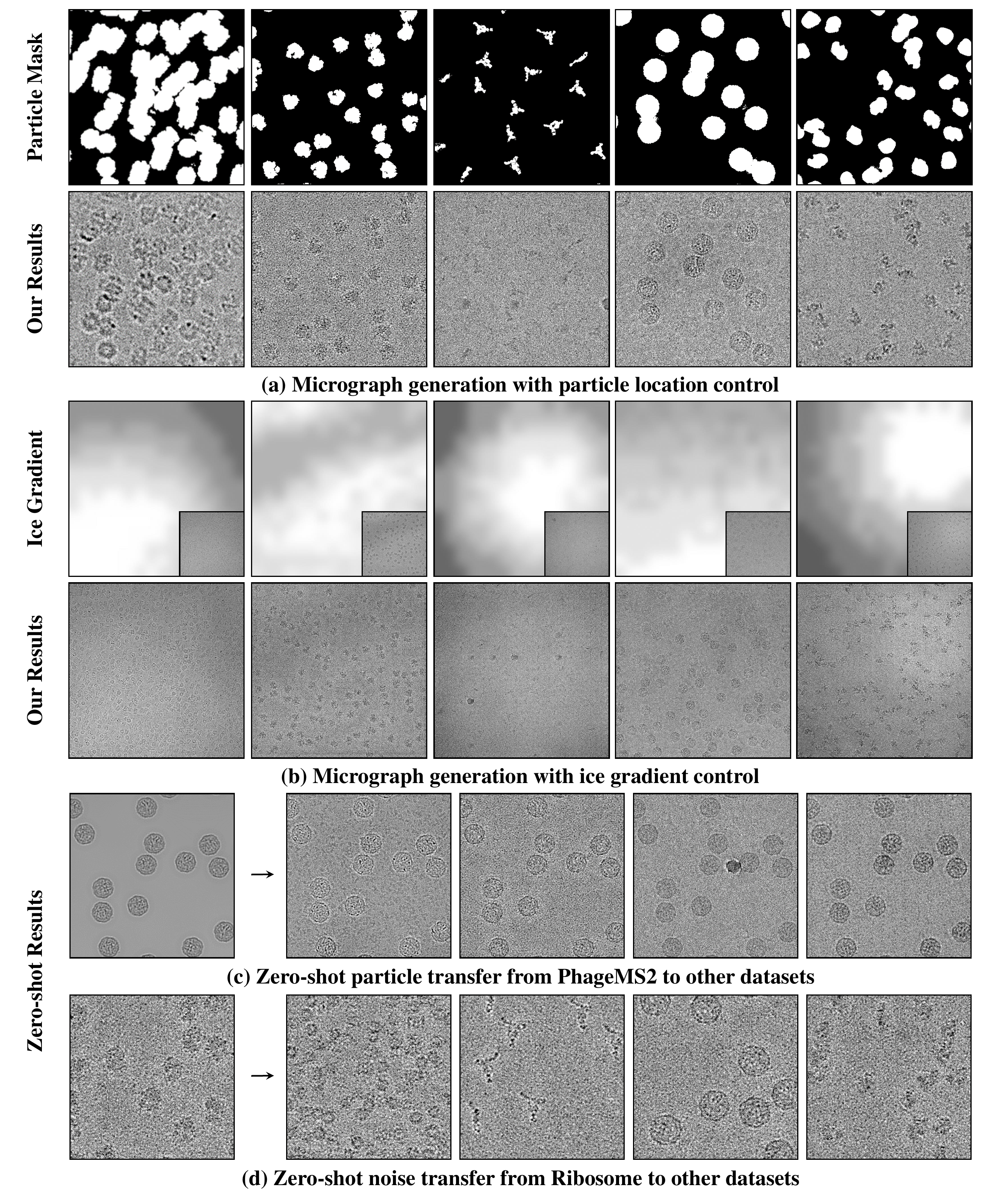}
\caption{
\textbf{(a) Location-controlled generation.}
\ours{} uses a particle mask to guide the sampling process in contrastive noise modeling. This method allows for precise control over the placement of particles in synthetic micrographs. \textbf{(b) Ice-gradient generation.} \ours{} calculates the ice gradient in real micrographs, as indicated in the lower right corner of each ice gradient image. This calculated ice gradient is then incorporated into the final images. \textbf{(c, d) Zero-shot transfer between particle and noise.} We show that \ours{} can generate convincing results on previously unseen datasets without the need for additional training. This demonstrates \ours{}'s ability for zero-shot transfer between different types of particles and noise.} 
    \label{fig-sup:pos-ice-zeroshot}
\end{figure}

\begin{figure}[htbp]
  \centering
  \includegraphics[width=0.9\textwidth]{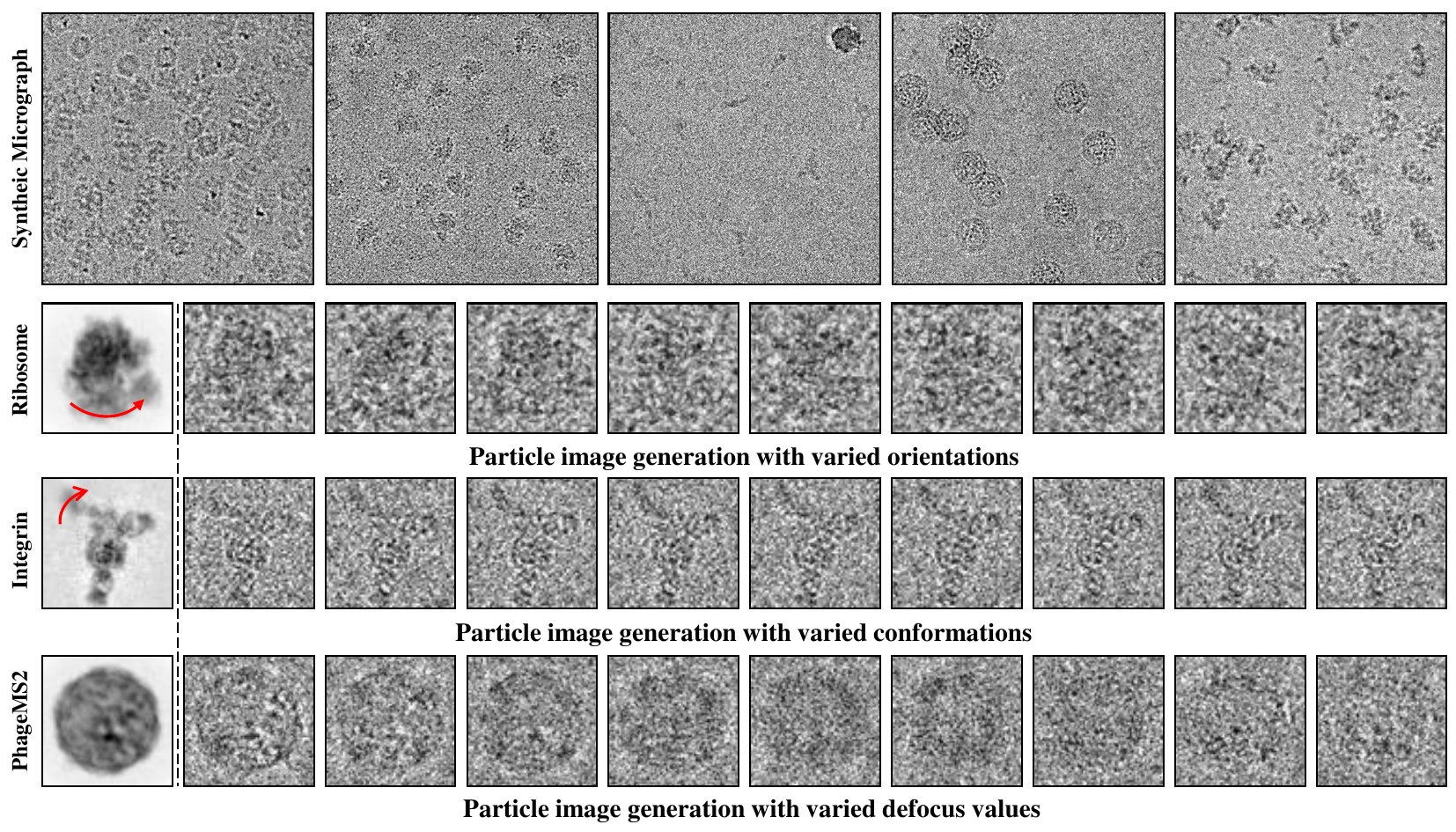}
   \caption{\textbf{Result gallery.} The first row showcases the diverse synthetic contents generated by \ours{} (from left to right for Proteasome, Ribosome, Integrin, PhageMS2, and HumanBAF). The second, third, and fourth rows demonstrate \ours{}'s ability to control the particle's pose, conformation, and defocus during the image generation. In every row, the leftmost column is the clean particle image, and the controlled variable is smoothly adjusted from left to right.
   }
   \label{fig-sup:gallery}
\end{figure}

We include supplementary results in Figure~\ref{fig-sup:pos-ice-zeroshot} and Figure~\ref{fig-sup:gallery}. These serve as an extension to Figure~\ref{fig:results} from our main paper. Here, we demonstrate \ours{}'s ability to generate synthetic micrographs with control over the particle mask, defocus value, and ice gradient. Additionally, we explore \ours{}'s zero-shot capability, showcasing its proficiency in generating authentic noise patterns on unseen datasets. These results show \ours{}'s potential in creating an extensive synthetic dataset in a wide range of particle and noise pattern combinations.
\begin{figure}[htbp]
    \centering
    \includegraphics[width=0.9\textwidth]{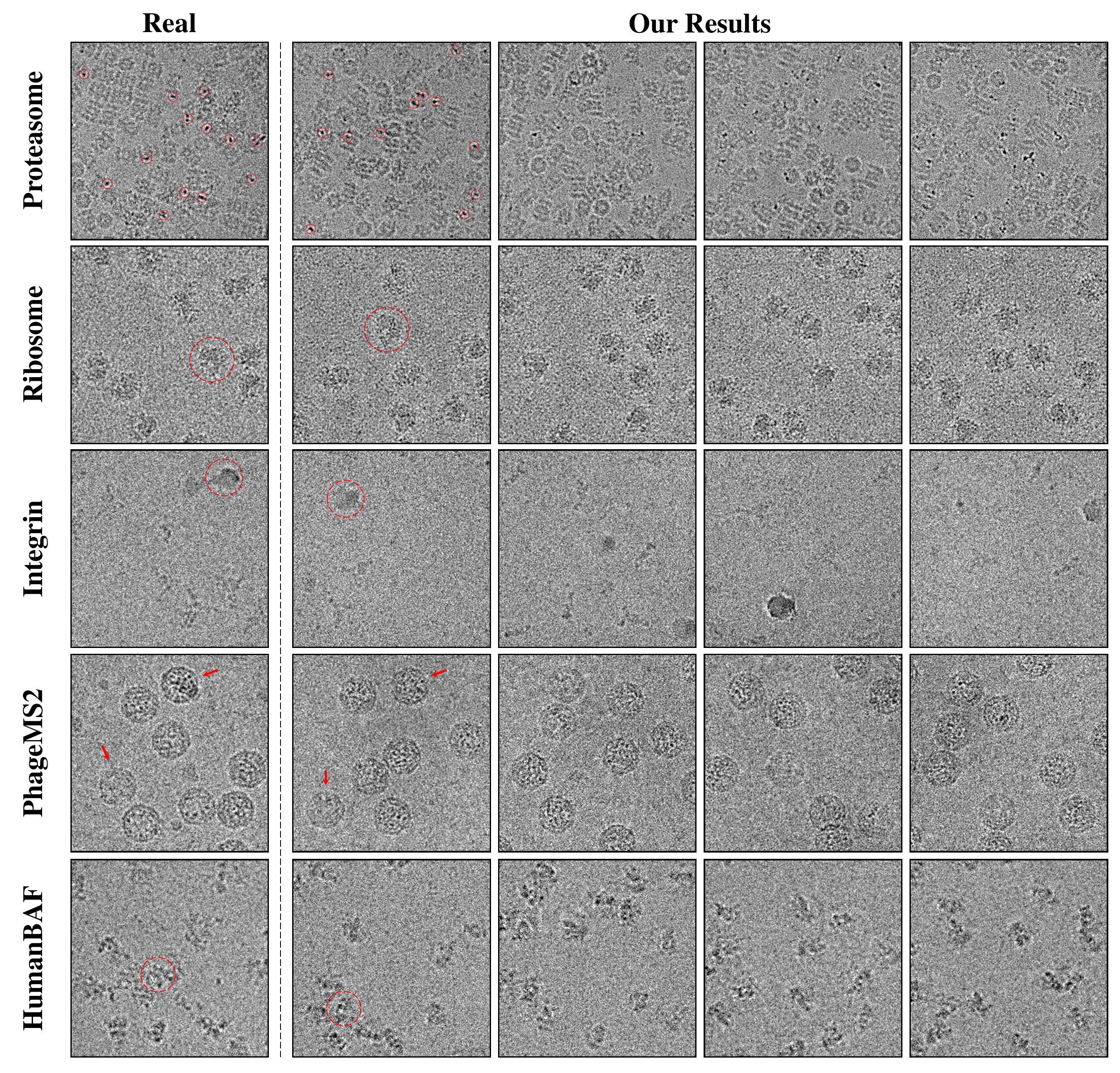}
    \caption{\textbf{Additional visual comparison with real datasets.}  Our evaluation of five complex real datasets demonstrates that \ours{} can accurately reproduce noise patterns, preserve structural details, and replicate specific anomalies like crystalline ice in Integrin. For each dataset, we highlight the visual characteristics adeptly captured by \ours{} with red circles or arrows.}
    \label{fig-sup:results}
\end{figure}

\section{Dataset Details}\label{app:data-pre}
We evaluate \ours{} across five diverse cryo-EM datasets. We source the \textbf{Proteasome}, \textbf{Ribosome}, and \textbf{Integrin} datasets from the Electron Microscopy Public Image Archive (EMPIAR)~\cite{iudin2023empiar}, a global public resource offering raw cryo-EM micrographs and selected particle stacks for constructing high-resolution 3D molecular maps. The \textbf{PhageMS2} and \textbf{HumanBAF} datasets were downloaded from the cryoPPP dataset~\cite{cryoppp} with provided manual particle picking annotations. 

\paragraph{Proteasome}
Thermoplasma acidophilum 20S proteasome~\cite{Campbell201528R} (EMPAIR-10025) contains 196 real micrographs and 49,954 manually filtered particles to achieve a high-resolution result at 2.8 \AA. The T20S proteasome forms a 700kDa complex that contains 14 $\alpha$-helices and 14 $\beta$-sheets organized by D7 symmetry. The captured micrographs comprise particles with high symmetry, random rotational orientations, and mutual occlusions.

\paragraph{Ribosome}
Human 80S ribosome~\cite{Wong2014CryoEMSO} (EMPIAR-10028) contains 1,081 real micrographs and 105,417 manually filtered particles to achieve a high-resolution result at 3.6 \AA. It is a large assembly, incorporating numerous RNA chains and proteins.

\paragraph{Integrin}
Asymmetric $\alpha$V$\beta$8 integrin~\cite{Campbell2020CryoEMRI} (EMPAIR-10345) contains 1,644 real micrographs and 84,266 manually filtered particles to achieve a high-resolution result at 3.3 \AA. The $\alpha$V$\beta$8 integrin is a complex of the human $\alpha$V$\beta$8 ectodomain with porcine L-TGF-$\beta$1, showing significant movement in its leg in the captured micrographs.

\paragraph{PhageMS2}
Bacteriophage MS2~\cite{empiar10075} (EMPAIR-10075) contains 300 real micrographs and 12682 manually filtered particles~\cite{cryoppp} to achieve a high-resolution result at 8.7 \AA. PhageMS2 consists of 178 copies of the coat protein, a single copy of the A-protein, and the RNA genome. Note that its size is significantly larger than others, making it a challenge for the particle picking model.

\paragraph{HumanBAF}
Endogenous human BAF complex~\cite{humanbaf} (EMPAIR-10590) contains 300 real micrographs and 62493 manually filtered particles~\cite{cryoppp} to achieve a high-resolution result at 7.8 \AA. 

\begin{figure}[t!]
  \centering
    \begin{minipage}[t]{0.48\textwidth}
    \centering
     {\includegraphics[width=\linewidth]{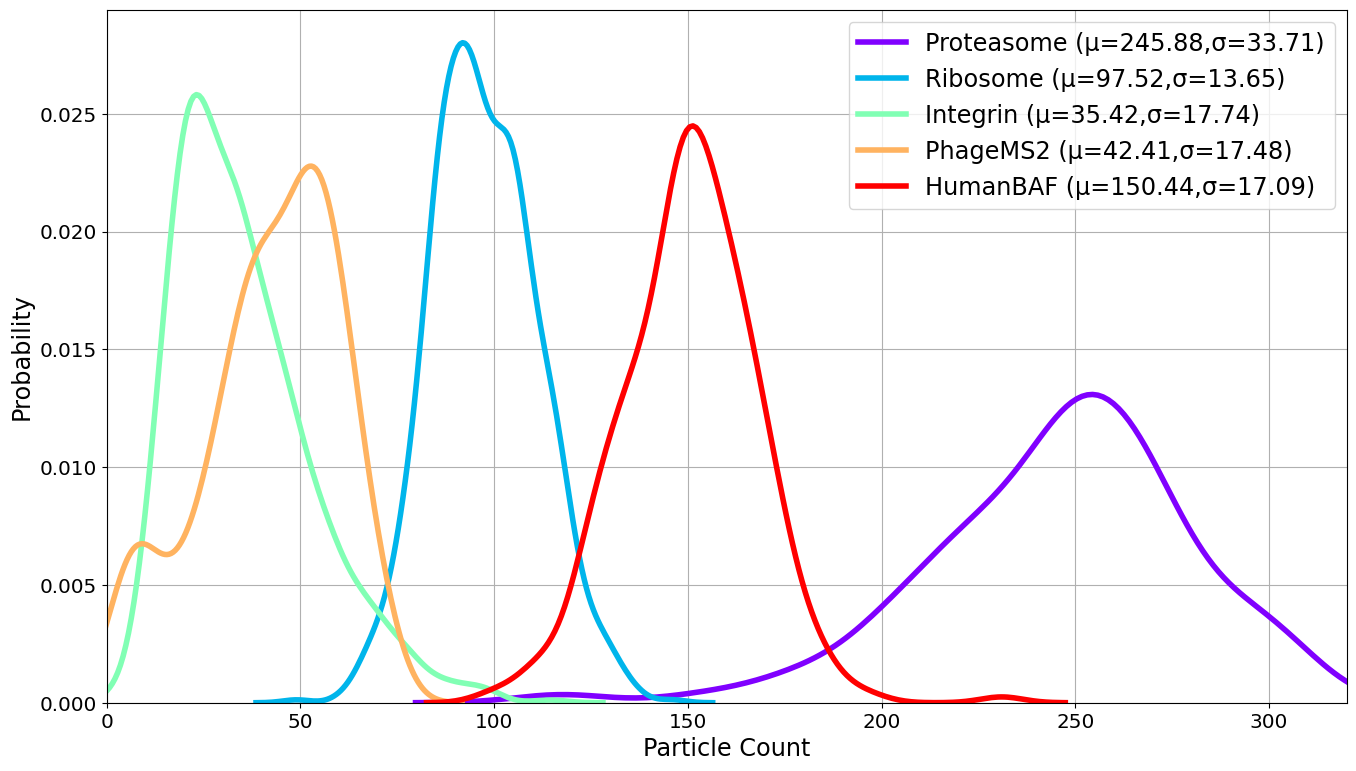}} 
    \caption{Probability curve of particle count in cryo-EM datasets.}
    \label{fig:particle_count}
  \end{minipage}
  \hfill
  \begin{minipage}[t]{0.48\textwidth}
    \centering
     {\includegraphics[width=\linewidth]{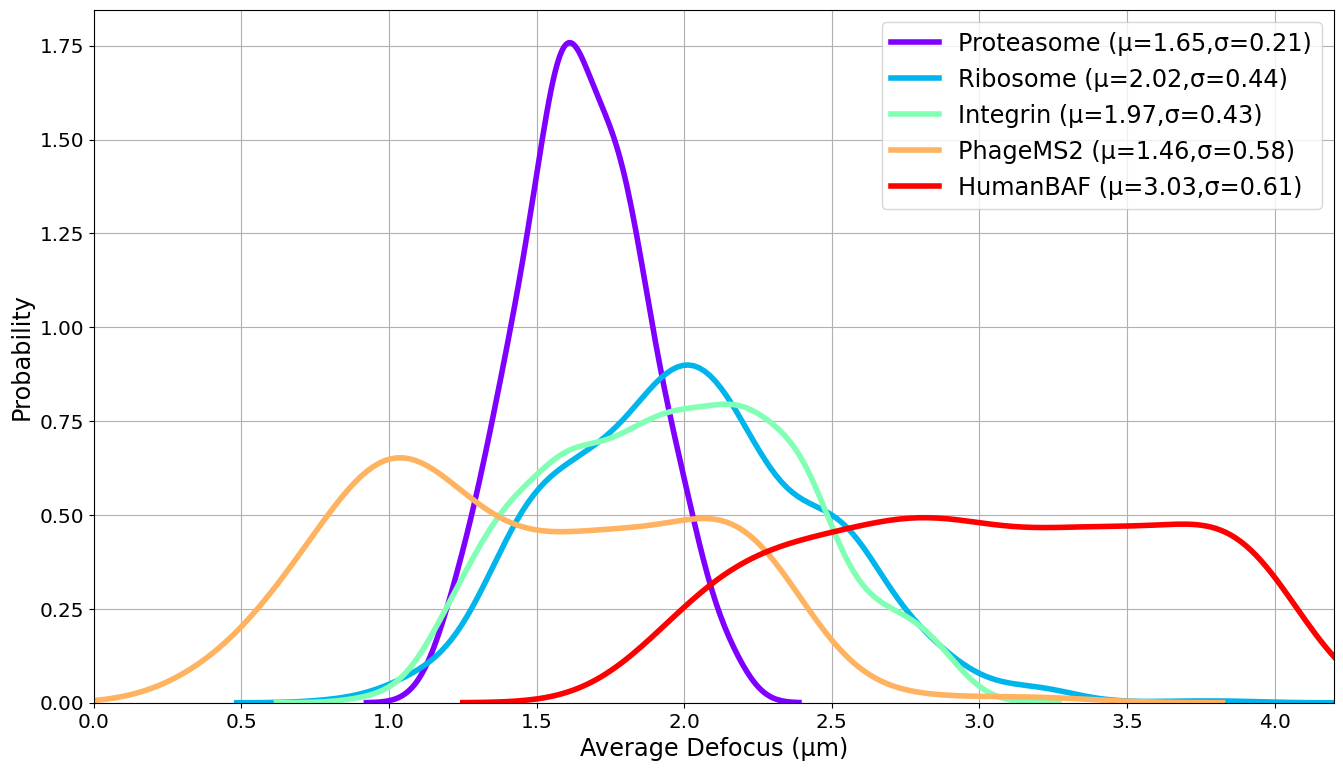}}
    \caption{Probability curve of defocus value in cryo-EM datasets.}
    \label{fig:defocus}
  \end{minipage}
\end{figure}

In Figure~\ref{fig-sup:results}, we demonstrate datasets in a wide variety of noise patterns, particle scales, and types. These characteristics present significant challenges to the accuracy of particle picking and pose estimation, which are critical for automated high-resolution cryo-EM reconstruction. Notably, \ours{} can generate similar visual effects such as the ring of defocus (Ribosome) and crystalline ice(Integrin).  We analyze the distribution of particle counts and defocus values across micrographs for each dataset, calculating their mean and standard deviation. Based on these analyses, we generate two Gaussian distributions to simulate particle count and defocus value accurately in \ours{}, as illustrated in Figures~\ref{fig:particle_count} and ~\ref{fig:defocus}. During the training phase, we select 100 micrographs from each real dataset as training samples for \ours{}, alongside other generative models like CUT~\cite{cut} and CycleGAN~\cite{cyclegan}. To ensure a fair comparison, we exclude these micrographs from our evaluation.

\paragraph{Dataset preprocessing.} 
To improve the quality of the training datasets, we meticulously remove "dirty" micrographs, which are defined as micrographs captured outside the intended target area, those exhibiting substantial jittering artifacts, and those only containing background ice. To improve the training efficiency, we downsample the images to $1024\times1024$ resolution. We enhance the contrast of the electron microscope images by adjusting the mean and standard deviation of the image intensity values to $150$ and $40$ (with a maximum value of 255), respectively. All these pre-processing steps are crucial for a stable and high-performance training process.

\paragraph{Initial 3D result.} 
We reconstruct a low-resolution 3D volume and simultaneously estimate the poses of input particle images using an \textit{ab}-initio reconstruction of cryoSPARC~\cite{cryosparc} with its default settings. To generate an initial neural volume with conformational changes, we employ an advanced neural method, cryoDRGN~\cite{cryodrgn} for continuous heterogeneous (dynamic) reconstruction, given the particle stack with estimated poses. We train the cryoDRGN 50 epochs by particles at a resolution aligned with the low-resolution 3D volume. Note that for homogeneous (static) datasets in our experiments, we directly utilize the 3D initial volume from cryoSPARC, as these datasets exhibit only negligible motions.

\paragraph{Ice gradient estimation.}
In Figure~\ref{fig:teaser} of the main paper, we show that ice gradients result in intensity variations across micrographs, leading to signal-to-noise ratio (SNR) disparities in local areas. Images from thicker specimen regions often exhibit a lower SNR than those from thinner regions. To mimic this physical effect, we employ IceBreaker~\cite{olek2022icebreaker}, an established method that addresses the challenge by treating the estimation of uneven ice thickness as a clustering problem.
We have developed a GPU-accelerated version of IceBreaker for a more efficient estimation of average brightness per class. By dividing the outcome by its maximum value, we achieve a normalized weight map. Then, to replicate the gradual variations in ice thickness, we apply a broad Gaussian kernel.

\paragraph{Optical aberration estimation.}
In cryo-EM, a high-energy electron beam interacts with the specimen, goes through a sophisticated optical imaging process, and is then captured by the detector as projection images. This optical modulation can be described by the Contrast Transfer Function (CTF), with the objective defocus being a particularly crucial parameter. Experimentally, biologists can modify the defocus to enhance contrast or capture higher-frequency signals. We directly incorporate CTF by applying it in our physical simulation process. Similar to CTFFIND4~\cite{rohou2015ctffind4}, CTF can be expressed as: 

\begin{align}
    \text{CTF}(w, \lambda, \textbf{g}, &\Delta f, C_s, \Delta\varphi) = \nonumber \\
    -\sqrt{1-w^2}&\sin[\chi(\lambda, |\textbf{g}|, \Delta f, C_s, \Delta \varphi)] \nonumber \\ 
     -w&\cos[\chi(\lambda, |\textbf{g}|, \Delta f, C_s, \Delta\varphi)],
     \label{equ:ctf}
\end{align}
where 
\begin{align}
    \chi\left(\lambda, |\textbf{g}|, \Delta f, C_s, \Delta \varphi\right) = \pi\lambda|\textbf{g}|^2\left(\Delta f-\frac{1}{2}\lambda^2|\textbf{g}|^2C_s\right)+\Delta\varphi
     \label{equ:ctf_2}.
\end{align}
$w $ is a relative phase contrast factor, \(\chi\) is the frequency-dependent phase shift function with the inputs including electron wavelength \(\lambda\), the spatial frequency vector \(\textbf{g}\), the objective defocus \(\Delta f\), the spherical aberration \(C_s\) and the phase shift \(\Delta\varphi\). \(w,\lambda, C_s, \Delta\varphi\) are the cryo-EM hardware parameters.

We utilize CTFFIND4 to estimate the range of the objective defocus, $\Delta f$. Subsequently, ResidualMFN~\cite{resmfn} is used to generate a specific 2D CTF image for any given defocus value. As depicted in Figure~\ref{fig:defocus}, we estimate the defocus value distribution in real cryo-EM datasets. For the physical simulation phase, defocus values are randomly chosen from simplified Gaussian distributions, based on previously calculated means and standard deviations.

\section{Evaluation Details}\label{app:baseline}
\subsection{Baseline Details}
\paragraph{Traditional noise simulations.} 
These baselines represent traditional simulation methods. The \textbf{Poisson} noise emulates the high-level noise caused by the stochastic nature of the detector under limited dosage conditions. Zero-mean \textbf{Gaussian} noise, as suggested by Vulovic et al.~\cite{Vulovic2013ImageFM}, is used to simulate a complex noise distribution, including readout noise, dark current noise, shot noise, and structural noise. We also combine these to create a \textbf{Poi-Gau} noise generation baseline. In all our experiments, we set the SNR to 0.1 for images generated by these baselines.

\paragraph{CycleGAN~\cite{CycleGAN2017, isola2017image}.} We use UNet~\cite{unet} as the generator and PatchGAN~\cite{patchgan} as the discriminator. The complete CycleGAN objective comprises five components: synthetic-to-real GAN loss, real-to-synthetic GAN loss, two-cycle consistency losses, and identity mapping loss. To maintain training stability, we assign their relative importance as 1, 1, 10, 10, and 0.5, respectively.

\paragraph{CUT~\cite{cut}.} We employ UNet~\cite{unet} as the generator and PatchGAN~\cite{patchgan} as the discriminator. The complete objective of CUT includes three components: synthetic-to-real GAN loss, NCE loss on synthetic and generated images, and NCE loss on real and real-identity images. We follow the original paper, setting their relative importance to 1, 1, and 1.

\paragraph{CycleDiffusion~\cite{cyclediff, improved-diffusion}.} CycleDiffusion necessitates the use of diffusion models for both the source and target domains. Consequently, we train two Denoising Diffusion Probabilistic Models (DDPMs) for each dataset: one for the synthetic image domain and one for the real image domain. Each model is trained on 100 images at a resolution of 512×512, with a batch size of 2, over 10,000 steps. For CycleDiffusion inference, we use the default settings. The associated code and models will be made available.

\begin{figure}[t]
    \centering
    \includegraphics[width=\textwidth]{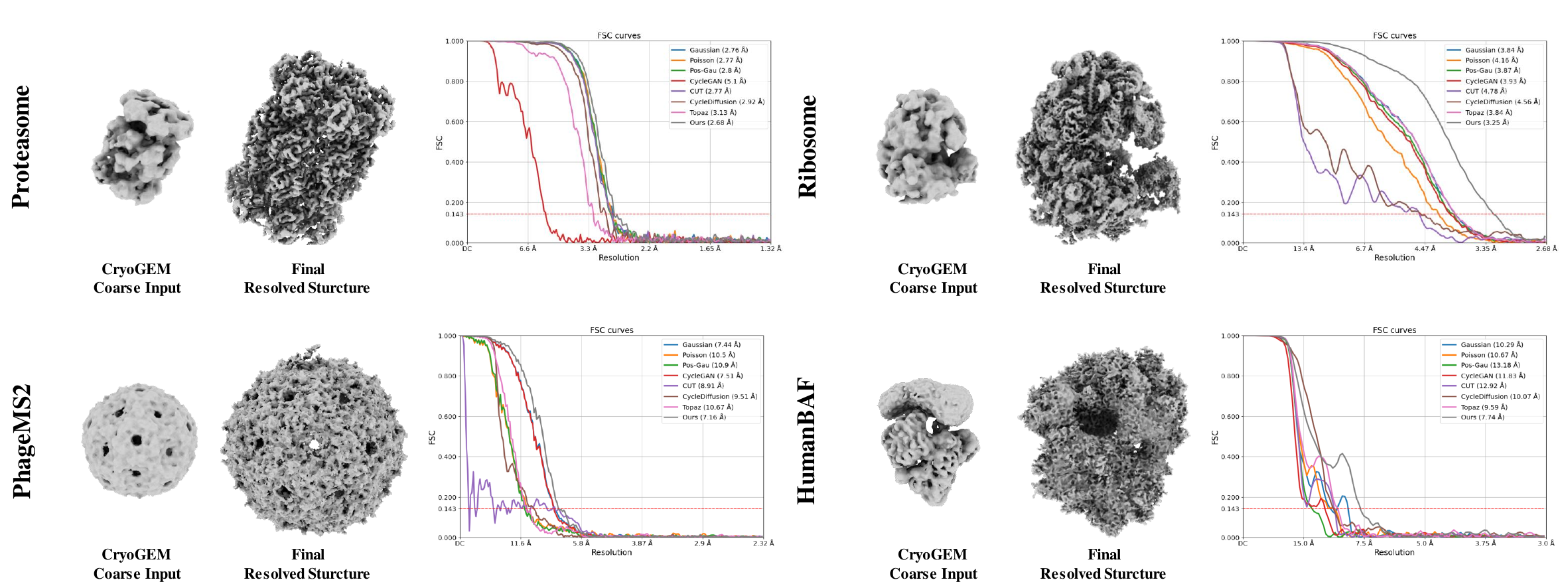}
    \caption{\textbf{Qualitative and Quantitative evaluation of \ours{}'s resolved structures and pose estimation.} We present cryo-EM resolved structures generated by \ours{} particle picking downstream workflow using the Proteasome, Ribosome, PhageMS2, and HumanBAF datasets. For each dataset, the coarse ab initio volume used for \ours{} physics-based module is shown on the left, and the final resolved structure with particles picked by \ours{}-finetuned Topaz is displayed in the middle. On the right, we provide the quantitative evaluation of particle picking reconstruction using the Fourier Shell Correlation (FSC)=0.143 metrics.
    }
    \label{fig-sup:pp-volumes}
\end{figure}

\subsection{Particle Picking Details} \label{app:pp-details}
Topaz~\cite{topaz} is a cryo-EM particle picking model that outputs the probability for each pixel belonging to a particle. With a threshold, it can filter out detection results. The training data for Topaz consists of complete micrographs and the central coordinates of particles within them. In practice, we select pre-trained ~\href{https://github.com/tbepler/topaz/blob/master/topaz/pretrained/detector/resnet8_u32.sav}{resnet8\_u32} as baseline \textbf{Topaz} model without fine-tuning. When we fine-tune Topaz by synthetic datasets from different baselines, we use 10 annotated micrographs to fine-tune the pre-trained Topaz for 20 epochs and then test it on the real evaluation datasets. Note that Topaz has already been pre-trained on the Proteasome and Ribosome datasets, which have been utilized in our experiments. Therefore, for these datasets, we infer the picking results at the pre-trained resolution of $512 \times 512$. For other datasets, we evaluate it at a resolution of $1024 \times 1024$, aligning with the fine-tuned resolution. 

\paragraph{Evaluation in terms of metric AUPRC.}
We retain the original particle picking results for comparison against the ground truth labels, without any thresholding. For a fair comparison of particle picking results, we prioritize our particle picks based on their confidence scores. Specifically, we select the highest-ranking 50,000 picks from each method for further filtering. For PhageMS2, however, due to the limited number of particles present in the micrographs, we only select the top 10,000 picks. By applying different thresholds divided into \(n\) intervals by \(\{\tau_i\}_{i=1}^{n-1}\), we can filter out varying prediction results and calculate corresponding precision and recall metrics against the true labels. Finally, we can compute the area under the precision-recall curve (AUPRC) as expressed by
 \begin{align}
    \sum_{k=1}^n \text{Pr}(k)(\text{Re}(k)-\text{Re}(k-1)),
    \label{equ:auprc}
\end{align}
where the precision \(\text{Pr}(k)\) represents the proportion of true positives in the prediction results with a probability greater than or equal to  \(\tau_k\), the recall \(\text{Re}(k)\) means the proportion of true positives in the prediction results with a probability greater than or equal to \(\tau_k\) out of the total number of true positives.

\paragraph{Evaluation in terms of metric Res(\AA ).}
We further assess the particle picking results by evaluating the final reconstruction resolution achieved using cryoSPARC~\cite{cryosparc}. The filtered picked results are fed into cryoSPARC for the subsequent reconstruction process. For static proteins such as Proteasome, Ribosome, PhageMS2, and HumanBAF, the reconstruction process involves a 1-class \textit{ab initio} reconstruction followed by homogeneous refinement. For dynamic proteins like Integrin, the reconstruction process includes a 5-class \textit{ab initio} reconstruction, heterogeneous refinement, and homogeneous refinement for each class. We select the best final resolution value as the reconstruction resolution. Importantly, since no 2D classification or particle filtering operations are performed before reconstruction, the resolution of the resolved structures directly reflects the precision of particle picking and the quality of the particles. This method ensures that no human bias is introduced.

\subsection{Pose Estimation Details}\label{app:pe-details}
To evaluate whether \ours{} can help with pose estimation tasks, we utilize the state-of-the-art \textit{ab}-initio reconstruction algorithm, cryoFIRE~\cite{cryofire}, from its official ~\href{https://github.com/ml-struct-bio/cryoFIRE}{Github page}. We employ synthetic datasets that provide a direct supervision mechanism through pairs of synthetic particle images and their corresponding true poses. In all experiments, we generate 100,000 particle images per baseline, each with a ground-truth orientation \(\{R_i\}_{i=1}^N\) and translation \(P=\{(R_i, T_i)\}_{i=1}^N\). To train the pose prediction module, we adopt the loss function for direct pose supervision by the loss introduced from ACE-EM~\cite{ace-em}.
\begin{align}
    \mathcal{L}_{\text{pose}}=\frac{1}{B}\sum_{i=1}^B\left[\frac{1}{9}\Vert R_i^{\text{gt}}-R_{i}^{\text{pred}}\Vert_F^2 + \frac{1}{2}\Vert T_{i}^{\text{gt}}-T_{i}^{\text{pred}}\Vert_1\right],
\end{align}
where B represents the training batch size, while \(R_i^{\text{gt}}\), \(R_{i}^{\text{pred}}\), \(T_{i}^{\text{gt}}\), and \(T_{i}^{\text{pred}}\) respectively represent the ground truth and network-predicted rotation and translation. We set B=256 in our experiments.
For various baselines, we train the pose estimation module for 200 epochs using only particles and their ground-truth poses. We then test the fine-tuned cryoFIRE's pose estimation module on the real datasets. Notably, we include the original \textbf{cryoFIRE} as an additional baseline in the pose estimation task. To determine the final reconstruction resolution, we randomly split the real data into two equal parts for separate reconstructions. The final reconstruction resolution is calculated by thresholding the Fourier Shell Curve (FSC) of the two reconstructions to 0.5.

\begin{figure}[t!]
    \centering
    \includegraphics[width=\textwidth]{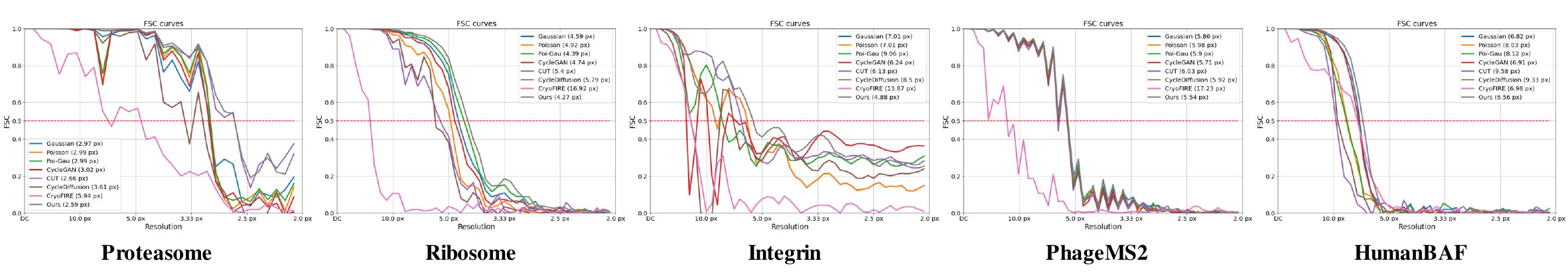}
    \caption{\textbf{Quantitative comparison of pose estimation using FSC curves.} Our approach demonstrates superior performance in the Res(px) metric. Notably, we present CryoFIRE, the sole neural-based reconstruction method that operates without ground truth particle poses.}
    \label{fig-sup:pe-fsc}
\end{figure}

\begin{figure}[t!]
    \centering
    \includegraphics[width=\textwidth]{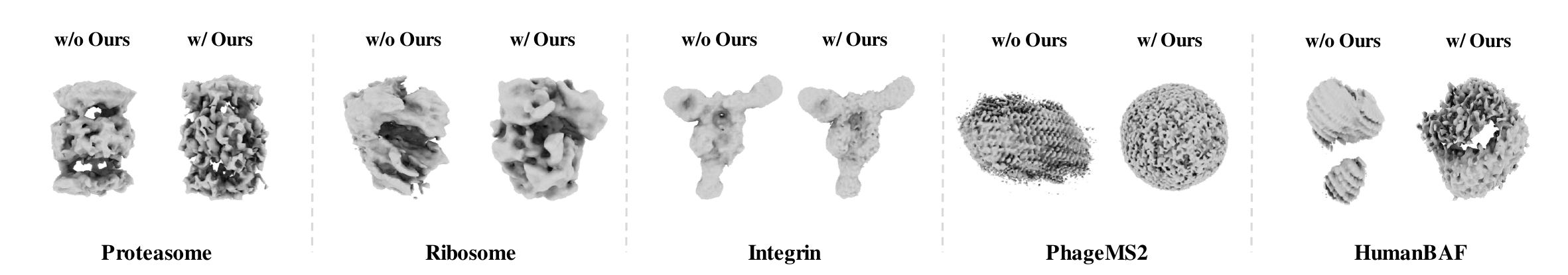}
    \caption{\textbf{Qualitative evaluation of performance improvement on pre-trained CryoFIRE.} The figure illustrates the detailed improvements in ab initio volumes. (Left: original CryoFIRE, Right: pre-trained CryoFIRE using \ours{}'s data).}
    \label{fig-sup:pe-improve}
\end{figure}

\paragraph{Evaluation in terms of metric Res(px).} We evaluate the pose estimation module on real datasets by using the estimated poses as training labels. For this purpose, we reconstruct two independent volumes from equally split particle stacks using the filter back-projection (FBP) method. We choose this traditional reconstruction algorithm over contemporary neural methods to directly assess pose accuracy without the bias of reconstruction algorithms. Due to the characteristics of FBP, some reconstructions' Fourier Shell Curves (FSC) do not meet the standard threshold of FSC=0.143. Therefore, we use a correlation coefficient of 0.5 for the Fourier shells as the metric Res(px) for pose estimation reconstruction resolution.

We showcase the performance of the original CryoFIRE, the only neural-based reconstruction method in our comparison that does not rely on ground truth particle poses. Figure~\ref{fig-sup:pe-fsc} displays the FSC curves for structures resolved during pose estimation. Additionally, we present the enhanced performance of pre-trained CryoFIRE when supplemented with \ours{}'s pose-labeled particle images, illustrating these improvements in Figure~\ref{fig-sup:pe-improve}. The formula for the Fourier Shell Correlation (FSC) is provided below:
\begin{equation}
    FSC(r)=\frac{\sum_{r_i\in r}F_1(r_i)\cdot F_2(r_i)^*}{\sqrt{\sum_{r_i\in r} \Vert F_1(r_i)\Vert^2 \cdot \sum_{r_i\in r} \Vert F_2(r_i)\Vert^2}}
    \label{equ:fsc}
\end{equation}
where \(F_1, F_2\) are the Fourier transforms of the FBP reconstructed volumes on equally split particle stacks, respectively. \(r\) represents all three-dimensional frequency components shown in a one-dimensional form. 

\paragraph{Evaluation in terms of metric Rot(rad).}
In addition to evaluating the resolution of resolved structures in pose estimation, we also report posing errors for all real images. For each dataset, we obtain the estimated ground truth pose from the cryoSPARC reconstruction pipeline. Due to a coordinate system misalignment issue, a rigid 6D-body alignment is applied to the poses predicted by the pose estimation module trained on generated data. This alignment ensures that the resolved structure is in the same coordinate system as the ground truth coarse volume. We then compute the mean angular error between the aligned estimated rotations $\hat{R}_i^{\text{pred}}$ and the ground truth rotations $R_i^{\text{gt}}$, in radians.
\begin{equation}
        \text{Rot(rad)}= \frac{180}{\pi n_{\text{rots}}}\sum_{i=1}^{n_{\text{rots}}}
        \arccos(\frac{R_i^{\text{gt}}\cdot v}{\Vert R_i^{\text{gt}}\cdot v \Vert}
        \cdot \frac{\hat R_i^{\text{pred}}\cdot v}{\Vert \hat R_i^{\text{pred}}\cdot v \Vert})
        \label{equ:rot-err}
\end{equation}

where $v$ represents a unit vector $(0, 0, 1)$.

\section{Societal Impacts}

The \ours{} technique, a novel physics-informed generative cryo-electron microscopy (cryo-EM) tool, has significant societal implications. we outline the potential positive and negative impacts, acknowledge uncertainties, and emphasize the broader implications of this technology.

\subsection{Positive Impacts}

\textbf{Improvement in biomedical fields}: By improving particle picking and pose estimation, \ours{} enhances 3D reconstructions of proteins, potentially accelerating the discovery and understanding of biomolecular structures, crucial for drug development and disease treatment.

\textbf{Reduction in labor-intensive tasks}: \ours{} automates the generation of annotated datasets, reducing the burden on human experts.

\subsection{Negative Impacts}

\textbf{Ethical and Misuse Concerns}: The ability to generate realistic synthetic data raises ethical concerns about misuse, such as manipulating research outcomes or creating misleading scientific evidence. Establishing guidelines for ethical use is crucial. There are several ways to prevent it from misusing including 1) developing and enforcing comprehensive ethical guidelines, 2) ensuring transparency in data generation, and forcing the users to claim the usage of \ours{} in their research projects.

\subsection{Conclusion}

CryoGEM offers transformative opportunities for cryo-EM and related fields, with significant benefits for biomedical research.  However, careful consideration of ethical implications, potential misuse, and long-term societal impacts is essential. By fostering a balanced and reflective approach, we can maximize the positive outcomes of \ours{} and mitigate potential risks, steering the technology in a beneficial direction for society.
 
\clearpage
\end{document}